\pdfoutput=1
\PassOptionsToPackage{svgnames}{xcolor}
\documentclass[10pt,a4paper,logo]{paper}

\usepackage[all]{hypcap}
\usepackage[authoryear, round, sort]{natbib}
\usepackage{array}
\usepackage{float}
\usepackage{wrapfig}
\usepackage{mathtools}
\usepackage{xspace}
\usepackage{afterpage}
\usepackage{fontawesome5}
\usepackage[most]{tcolorbox}

\definecolor{title_blue}{HTML}{204899}
\definecolor{cite_blue}{HTML}{044dc1}
\definecolor{cite_purple}{HTML}{7406a7}

\hypersetup{
    colorlinks = true,
    citecolor = {cite_blue},
    linkcolor = {cite_purple},
    urlcolor = {cite_purple},
}

\newcommand{\paragraphgap}{-8pt}
\let\origparagraph\paragraph
\makeatletter
\renewcommand{\paragraph}{\if@nobreak\else\vspace{\paragraphgap}\fi\origparagraph}
\makeatother


\let\cite\citep

\def\eg{\emph{e.g.,}\xspace}
\def\ie{\emph{i.e.,}\xspace}

\newcommand{\ourmodel}{FloorSAV\xspace}
\DeclareRobustCommand{\ourbench}{SAVED-Bench\xspace}

\newcommand{\gain}[1]{\,$_{\textcolor{PineGreen}{\text{+#1}}}$}
\newcommand{\loss}[1]{\,$_{\textcolor{Red}{\text{--#1}}}$}

\title{FloorSAV: Elucidating Spatial Audio-Visual Context with 2D Floormap for AV-LLMs}

\reportnumber{}
\newcommand{\headerdate}{Oct 2026}

\author[1]{Kyeong-Rae Kim}
\author[1,\footnotesize{$\dagger$}]{Sungnyun Kim}
\author[1,\footnotesize{$\dagger$}]{Tae-Hyun Oh}

\affil[1]{Korea Advanced Institute of Science and Technology}

\authornote{\textsuperscript{$\dagger$}Corresponding authors}

\newcommand{\projecturl}{https://byulharang.github.io/FloorSAV/}
\newcommand{\contactblock}{{\Affilfont
  \faEnvelope[regular]~\texttt{\{kimkyeongrae, ksn4397, taehyun.oh\}@kaist.ac.kr}\\[1pt]
  \faGlobe~\href{\projecturl}{\textcolor{cite_blue}{\texttt{\projecturl}}} }}
\correspondingauthor{}

\begin{abstract}

While 3D spatial reasoning in dynamic egocentric environments is crucial for embodied intelligence, audio-visual large language models (AV-LLMs) lack explicit mechanisms to process and internalize global geometry directly from raw sensory streams. Existing approaches either require costly fine-tuning or underutilize the model's cross-modal reasoning capacities. In this paper, we propose \textbf{\ourmodel}, a novel framework that explicitly grounds spatial audio-visual context by rendering a dynamic \textbf{2D floormap}. By integrating 3D point clouds, camera trajectories, spatial audio cues, and semantically grounded object landmarks, we inject this floormap into the AV-LLM as a synchronized stream with an egocentric video. AV-LLMs utilize their multi-modal capabilities to jointly reason over visual, auditory, and geometric cues in a single inference with floormap interpretation guidance. We further introduce \textbf{\ourbench} (Spatial Audio-Visual Egocentric Benchmark with Dynamic Agents), constructing essential tasks of spatial capability in real-world scenarios: \emph{dynamic relativity}, \emph{regional}, and \emph{path reasoning} QAs. \ourmodel improves AV-LLMs' spatial reasoning on various tasks from both \ourbench and SAVVY-Bench. Studies with ground-truth floormaps demonstrate the substantial potential of \ourmodel with accurate spatial information.

\end{abstract}

\begin{document}

\maketitle

\par\medskip\noindent
\begin{minipage}{\linewidth}
    \captionsetup{type=figure}
    \centering
    \includegraphics[width=\linewidth]{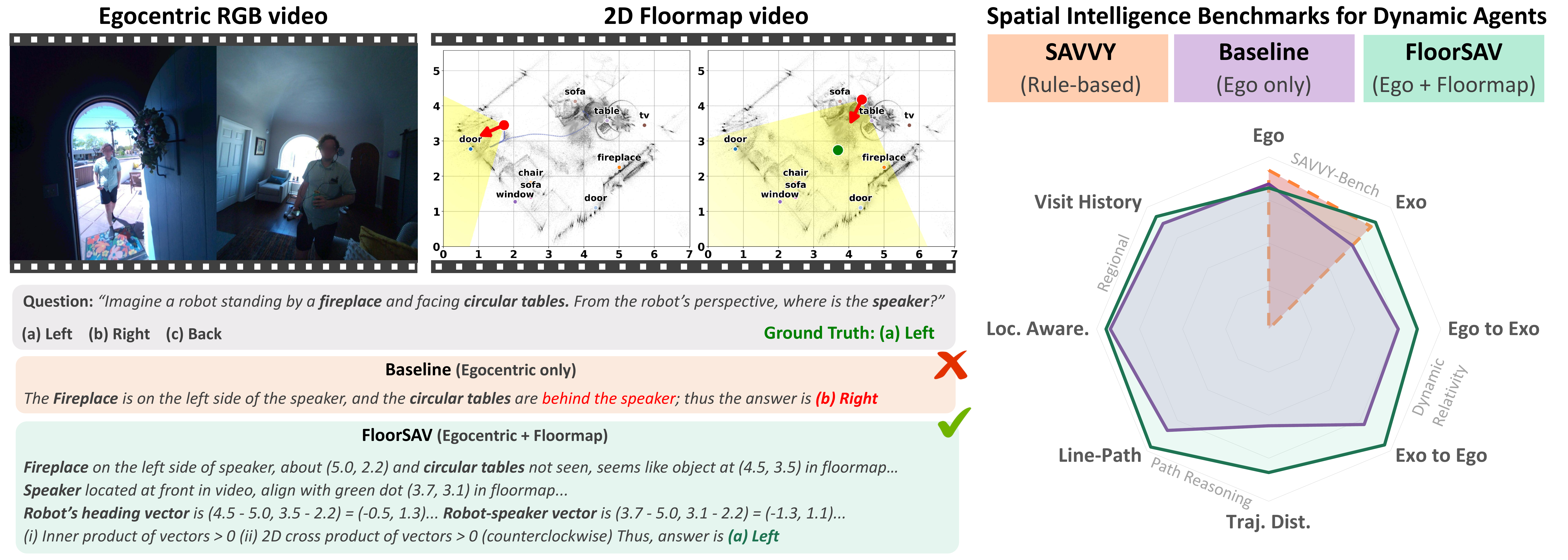}
    \captionsetup{font=footnotesize}
    \vspace{-15pt}
    \caption{\textbf{\ourmodel improves spatial reasoning, and our \ourbench widely complements SAVVY-Bench.} \textit{\textbf{(Left)}} \ourmodel enables MLLMs to jointly reason over the egocentric and 2D floormap videos to resolve various spatial tasks. \textit{\textbf{(Right)}} On the comprehensive evaluation including \ourbench, which comprises cross-agent, region-aware, and environment-aware tasks, \ourmodel improves the egocentric video-only baselines, with Gemini series \cite{comanici2025gemini, google2026gemini36}, for most of the spatial QA tasks.}
    \label{fig:intro}
    \label{fig:stage_radar}
\end{minipage}
\par\medskip

\afterpage{%
\begin{figure}[!t]
    \centering
    \includegraphics[width=\linewidth]{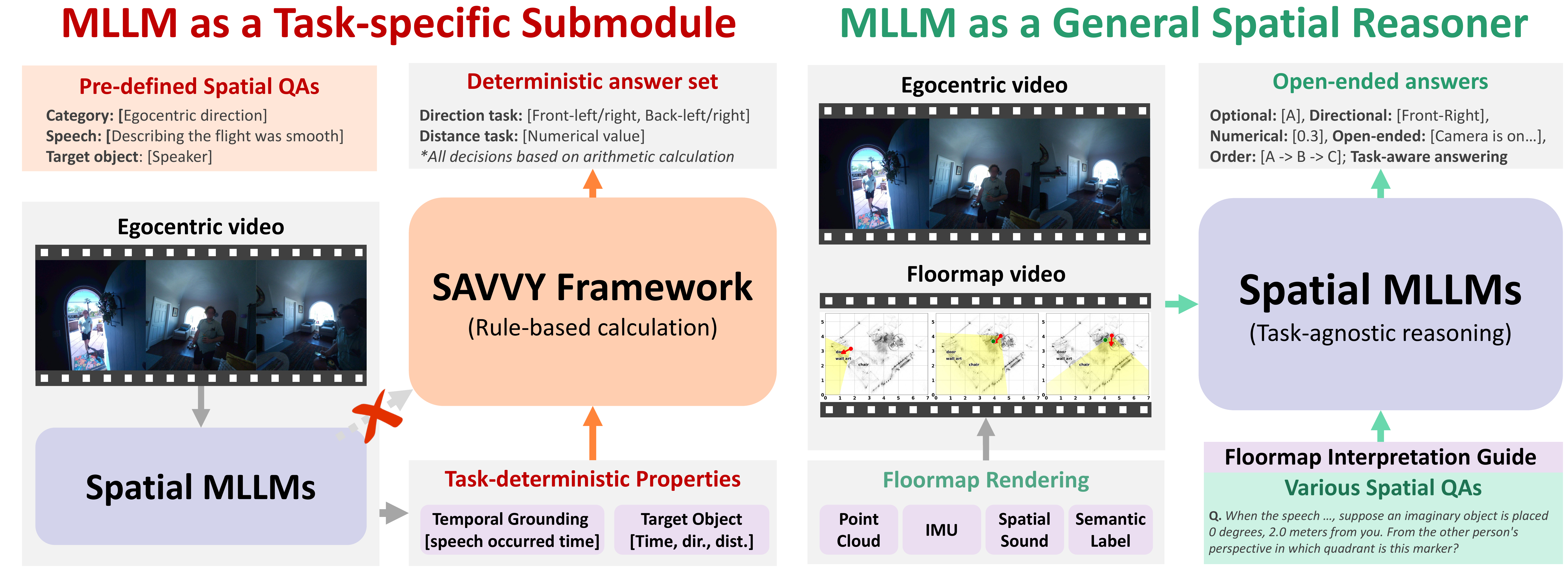}
    \captionsetup{font=footnotesize}
    \vspace{-15pt}
    \caption{\textbf{Comparison of SAVVY and \ourmodel frameworks.} SAVVY underutilizes MLLMs as a submodule, relying on pre-defined QA tasks and computing answers by a deterministic process. In contrast, \ourmodel integrates geometry, agents, sound estimates, and key objects into a floormap, then injects it with synchronized egocentric video to the MLLM, where it jointly reasons over both streams to answer various spatial tasks.}
    \label{fig:framework_comparison}
\end{figure}
}

\section{Introduction}
\label{sec:intro}

Egocentric audio-visual understanding~\cite{grauman2022ego4d, huang2023egocentric, seth2026egoavu, zhu2026egosound} is central to embodied intelligence. Systems operating from a first-person perspective must jointly process visual and acoustic data alongside motion to answer complex spatial questions, such as identifying the direction of a speaker or estimating distances of sound sources~\cite{yun2021pano, li2022learning, yang2022avqa}. However, comprehensive 3D spatial reasoning~\cite{zheng2025multimodal}, the cognitive ability to navigate changing environments by tracking dynamic objects and transforming first-person observations into consistent global representations, remains largely unsolved. As demonstrated by recent benchmarks like SAVVY-Bench~\cite{chen2025savvy}, even the most advanced audio-visual large language models (AV-LLMs)~\cite{cheng2024videollama, geng2025longvale, ye2025omnivinci} fall significantly below human-level performance on fine-grained spatial tasks.

This performance gap stems from the inherent limitations of relying on raw egocentric streams. Visual reasoning suffers from severe spatial bias due to the continuously moving, narrow field-of-view (FoV) of wearable cameras, making it exceptionally difficult for models to internalize global 3D geometry from video alone~\cite{yang2025thinking}. Spatial audio further compounds this difficulty; tracking transient sound sources from moving microphones is inherently noisy~\cite{diaz2020robust, dementyev2026phasecoder, sridhar2025spatial}, and current AV-LLMs typically process downmixed monaural audio that discards critical spatial attributes~\cite{gong2024listen, chu2023qwen, sun2024video, xu2025qwen3}. Consequently, models struggle to ground dynamic audio-visual events into a coherent physical space.

Recent efforts to address these challenges generally fall into two categories, both facing significant hurdles. Training-based methods~\cite{zheng2024bat, biswas2025owl, liu2026jaeger} attach fine-tuned spatial encoders to AV-LLMs, which requires access to proprietary model weights and expensive domain-specific 3D supervision. Conversely, training-free pipelines, as exemplified by SAVVY~\cite{chen2025savvy}, avoid architectural modifications by extracting spatial information externally. However, these approaches relegate the AV-LLM to simple preliminary sub-tasks like temporal localization, while fine-grained spatial tracking and global map construction are executed by offline, rule-based modules. Hence, the AV-LLMs do not jointly process the spatial audio-visual context, limiting its capability for addressing
diverse open-ended tasks and leaving its powerful cross-modal reasoning capacity underutilized.

A top-down map suits this setting. It compresses a long, narrow-FoV video into a single metric reference frame whose coordinates the model can read off directly. In language models, the map has been shown as a form of external visualization to aid spatial reasoning~\cite{wu2024mind}, although current models remain well below human accuracy on top-view tasks~\cite{li2024topviewrs}. Rendering the map per frame also keeps it synchronized with the video, so a moving camera and moving sound sources appear as visual motion rather than as text the model has to integrate.

Inspired by this, we introduce \textbf{\ourmodel}, a task-agnostic training-free framework that elucidates spatial context for AV-LLMs through an explicit visual representation (Figure\,\ref{fig:intro}). Using the Aria Everyday Activities (AEA) dataset~\cite{lv2024aria}, we construct a per-frame \textbf{2D floormap} dynamically annotated with camera trajectories and estimated sound source positions. By feeding this floormap as a second synchronized video stream alongside the egocentric view, the model can jointly reason over visual appearance, auditory cues, and global scene geometry in a single inference on spatial tasks.

We further introduce \textbf{\ourbench} (\underline{\textbf{S}}patial \underline{\textbf{A}}udio-\underline{\textbf{V}}isual \underline{\textbf{E}}gocentric Benchmark with \underline{\textbf{D}}ynamic Agents), which broadly evaluates spatial intelligence for dynamic agents in the real world. %
While recent egocentric spatial benchmarks~\cite{yang2025thinking, du2024embspatial} evaluate broad spatial tasks, they lack a crucial element: dynamic agents moving through the scene. Although SAVVY-Bench~\cite{chen2025savvy} incorporates dynamic agents, its evaluation is limited to direction and distance estimation within a given reference frame. \ourbench introduces broader spatial reasoning tasks relevant to dynamic agents' everyday interactions through three complementary categories: \textit{dynamic relativity}, \textit{regional}, and \textit{path reasoning} QAs.
These assess cross-agent, semantic region-aware, and path-aware cognition abilities, respectively.
Evaluations on \ourbench demonstrate that our \ourmodel strategy improves spatial reasoning by jointly utilizing multi-sensory data and generalizes to various spatial tasks that are inherently intractable for SAVVY (Figure\,\ref{fig:intro}). Our main contributions are as follows:
\begin{itemize}[leftmargin=*, itemsep=2pt, topsep=0pt]
    \item \textbf{Floormap rendering and guidance.} \ourmodel integrates multi-sensory data and key objects onto 2D floormaps as explicit representations. The floormap guide helps AV-LLMs jointly reason over both egocentric and 2D floormap videos.

    \item \textbf{Broader spatial reasoning benchmark.} \ourbench, which comprises \emph{dynamic relativity}, \emph{regional}, and \emph{path reasoning} QAs, comprehensively spans cross-agent, region-aware, and path-aware tasks to evaluate essential spatial capabilities for dynamic agents in real-world scenarios.

    \item \textbf{\ourmodel improves spatial intelligence across benchmarks.} \ourmodel shows benefits on both \ourbench and SAVVY-Bench. Compared to the egocentric view-only baseline (\ie Gemini), \ourmodel improves the overall score, from 58.3\% to 65.3\% on \ourbench, and from 50.9\% to 54.7\% on SAVVY-Bench. Further studies with ground-truth agent positions and objects show the potential of \ourmodel with an overall score of 71.3\% on \ourbench.
\end{itemize}

\section{Related Work}

\subsection{Spatial Understanding in MLLMs}

To endow MLLMs with 3D spatial reasoning for embodied AI~\cite{hong20233d, huang2024embodied, zheng2025video, deng20253d,yu2026splatreasoner}, early vision-based approaches have focused on spatial QA and visual grounding~\cite{chen2023shikra, chen2024spatialvlm, peng2024grounding}. However, they inherently suffer from visual bias and frequently hallucinate in occluded or out-of-view regions~\cite{li2023evaluating, yu2023mm, yang2025thinking}. To mitigate this, audio-based MLLMs~\cite{gong2024listen, chu2023qwen} utilize acoustic cues via multi-microphone arrays for estimating distance and direction~\cite{zheng2024bat, you2026world}, often enhanced by Chain-of-Thought (CoT) reasoning~\cite{biswas2025owl} or microphone-agnostic encoders~\cite{dementyev2026phasecoder}. Nevertheless, these audio-centric approaches discard rich visual semantics and heavily rely on simulated data, leaving their real-world applicability unexplored. Consequently, to exploit cross-modal complementarity, unified AV-LLMs have emerged~\cite{zhang2023video, cheng2024videollama, ye2024cat, ye2025omnivinci, zhong2026omni}. Progress here, however, is largely measured on audio-visual QA benchmarks that ask what and when rather than where~\cite{li2022learning, yang2022avqa}, with spatial grounding appearing only in restricted settings such as 360$^\circ$ video~\cite{yun2021pano}. For instance, recent works jointly fine-tune audio-visual projectors and LLMs to solve spatial QA tasks~\cite{ryu2026here}. However, despite first-person perception being a fundamental prerequisite for dynamic, embodied intelligence, there remains a critical lack of research explicitly leveraging AV-LLMs for egocentric spatial reasoning.

\subsection{Egocentric Audio-Visual Spatial Reasoning}

Egocentric vision is crucial for embodied intelligence but suffers from restricted FoV, driving the need for audio-visual integration and multi-sensory benchmarks~\cite{grauman2022ego4d, mangalam2023egoschema, plizzari2025omnia, zhu2026egosound, seth2026egoavu}. To enhance spatial awareness, JAEGER~\cite{liu2026jaeger} fine-tunes audio-visual encoders, but faces a sim-to-real gap and incompatibility with proprietary LLMs. Alternatively, SAVVY~\cite{chen2025savvy} introduces a training-free pipeline using a real-world dataset~\cite{lv2024aria}, whose synchronized point clouds, 6-DoF poses, and multi-channel audio also underpin recent egocentric 3D benchmarks~\cite{straub2024efm3d}. However, it underutilizes the MLLM's powerful cross-modal capabilities, and SAVVY-Bench only scopes narrow spatial relations within a single dynamic agent's viewpoint. In contrast, \ourmodel endows MLLMs with explicit spatial reasoning by incorporating a multi-sensory integrated 2D floormap. We further suggest a comprehensive benchmark that evaluates essential spatial capabilities. This aligns with a growing consensus in vision-language models that providing explicit spatial abstractions~\cite{jatavallabhula2023conceptfusion, gu2024conceptgraphs,jun2025dr}, whether through native 3D coordinate grounding~\cite{wang2025n3d}, mental scene simulations for exocentric (allocentric) perspective shifts~\cite{lee2025perspective}, or 2D maps for navigation and top-view reasoning~\cite{huang2023visual, wang2023gridmm, chen2026floorplan, li2024topviewrs}, is essential to unlock advanced spatial reasoning capabilities~\cite{wu2024mind}.

\section{Method: \ourmodel}
\label{sec:method}

Spatial reasoning in dynamic 3D environments is fundamentally challenging for AV-LLMs when relying solely on egocentric sensory streams, as a narrow FoV obscures global geometry and transient audio events are difficult to track.
To address this limitation, \textbf{\ourmodel} elucidates the spatial audio-visual context through an explicit \textbf{2D floormap}. Unlike SAVVY~\cite{chen2025savvy}, which relegates the AV-LLM to preliminary sub-tasks followed by offline rule-based mapping, \ourmodel renders the global map and injects it alongside the corresponding egocentric video into the AV-LLM, thus unlocking its cross-modal reasoning capacity (Figure\,\ref{fig:framework_comparison}).

\subsection{Floormap Rendering via Explicit Spatial Grounding}
\label{subsec:grounding}

We render a compact top-down 2D floormap from the spatial audio-visual signals in the AEA dataset~\cite{lv2024aria} in three steps (details in Appendix\,\ref{app:floormap_rendering}).

\paragraph{Step 1: Base map and acoustic tracking.}
We first project 3D point clouds into a 2D metric grid with one-meter ticks to establish the static room layout and boundaries.
The camera wearer's pose (position and orientation) is taken from the dataset, and the distance and direction-of-arrival (DoA) of sound sources are estimated from the multi-channel audio~\cite{dibiase2000high, schwarz2015coherent, diaz2020robust}, all projected onto the base map.

\paragraph{Step 2: Visual grounding via spatial markers.}
To make the map intuitive for the model, we draw these estimates with explicit visual markers: a red dot/arrow and a yellow wedge for the camera's current position/heading and FoV, a green dot for the estimated sound source, and a continuous blue polyline for the camera's historical trajectory.

\begin{figure}[t]
    \centering
    \includegraphics[width=\linewidth]{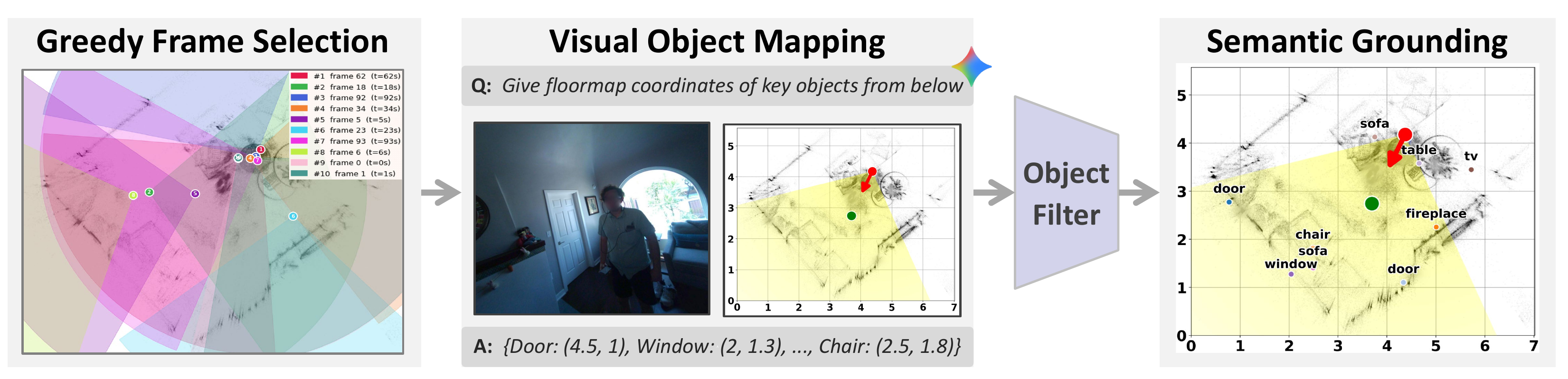}
    \captionsetup{font=footnotesize}
    \vspace{-15pt}
    \caption{\textbf{Semantic grounding process.} Semantically enriched floormap is generated by a pipeline of keyframe selection, mapping visual objects via VLMs, and filtering, providing semantic context for better reasoning.}
    \label{fig:semantic_grounding}
\end{figure}

\paragraph{Step 3: Semantic grounding via VLM.}
However, the markers do not have information of spatial relations with the environment, so we enrich the floormap with key object landmarks (Figure\,\ref{fig:semantic_grounding}) following this pipeline: (1) keyframes are selected by greedy maximum coverage of the map, (2) an off-the-shelf VLM (\eg Gemini~\cite{comanici2025gemini}) detects static objects in each keyframe and extracts their spatial coordinates, and (3) the detections are normalized, filtered, and merged across frames into labels overlaid on the map as text.

\subsection{Floormap-Aware Inference}
\label{subsec:inference}

Once the 2D floormap frames are constructed, we assemble them into a video synchronized with the egocentric footage. Both egocentric and floormap videos are given to the AV-LLM via a dual-stream video injection strategy with a brief interpretation guide.

\paragraph{Dual-stream injection.}
We feed the egocentric and floormap videos as two separate, synchronized streams sampled at the same frame rate, rather than as a single side-by-side video~\cite{chen2026floorplan} or a static map image. We empirically found that the side-by-side concatenation introduces severe prediction biases toward the side of the frame that the egocentric view occupies (Appendix\,\ref{app:prompting_strategies}), while a static image loses the motion of the camera and the sound source. The dual-stream video injection preserves the original geometric layout intact.

\paragraph{Floormap guide.}
We construct a guide that explains the rendered marks in the floormap: colored dots for the observer's position and detected sound sources, camera FoV cues, camera trajectories, and text labels for key objects. It helps the model use the egocentric view as the primary reference and refer to the floormap when spatial details are uncertain (full prompts in Appendix\,\ref{app:prompts}).

\section{Spatial Reasoning Benchmarks}
\label{sec:Spatial Reasoning Benchmarks}
\label{sec:ourbench}

\begin{figure}[!t]
    \centering
    \includegraphics[width=\linewidth]{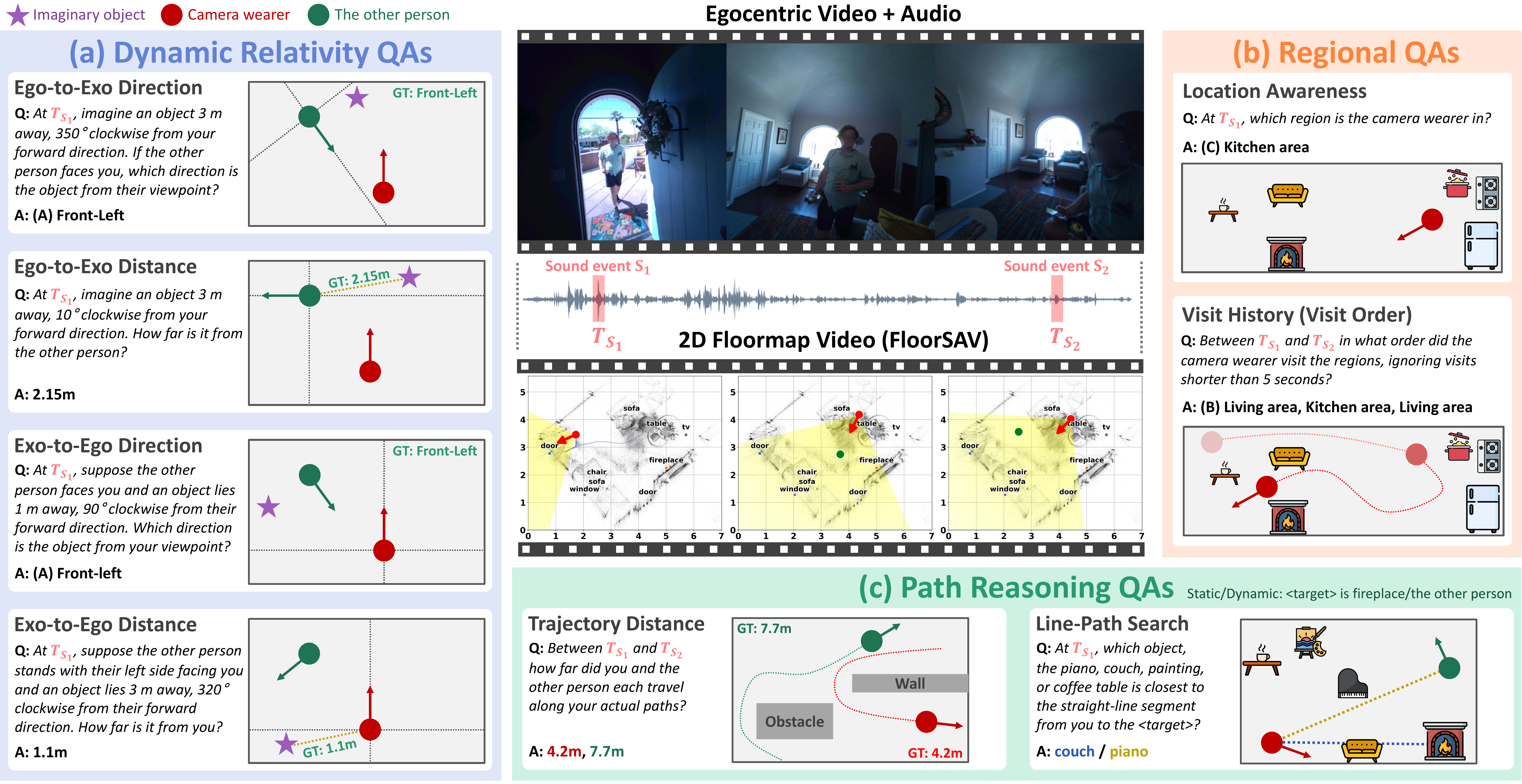}
    \captionsetup{font=footnotesize}
    \vspace{-15pt}
    \caption{\textbf{\ourbench at a glance.} Illustrative question-answer pairs cover \textbf{(a)} dynamic relativity, \textbf{(b)} regional, and \textbf{(c)} path reasoning tasks. Sound events ($S_t$) identify the relevant moments ($T_{S_t}$) or intervals, while synchronized egocentric and floormap views provide spatial context.}
    \label{fig:stage_task_guide}
    \label{fig:saved_at_a_glance}
\end{figure}

\subsection{SAVVY-Bench}
\label{subsec:savvy_bench}

SAVVY-Bench~\cite{chen2025savvy} is the first benchmark for 3D spatial reasoning in audio-visual egocentric scenes, sourced from the AEA dataset~\cite{lv2024aria} which records daily-life scenarios. Its tasks are separated by reference viewpoint and ask about direction or distance at the moment when the target sound occurred.
\textbf{Egocentric (Ego)} utilizes the viewpoint of the recorded video (camera wearer), and \textbf{Exocentric (Exo)} assumes a hypothetical observer (\eg robot) specified in the question.

\subsection{\ourbench: Towards Dynamic Agents in Real-World Scenarios}
\label{subsec:SAVED-Bench: Spatial Audio-Visual Egocentric Benchmark with Dynamic Agents}
Although SAVVY-Bench provides spatial reasoning tasks involving interactive agents, each question is evaluated from a single fixed viewpoint at the moment of a target speech event and is limited in scope to the direction/distance estimation. However, spatial intelligence for agents in real-world scenarios requires broader aspects of spatial reasoning~\cite{yang2025thinking, wang2025site}.
Therefore, we introduce \textbf{\ourbench} to evaluate spatial reasoning involving dynamic agents in real-world scenarios. The benchmark comprises three complementary categories that target cross-agent, region-aware, and path-aware reasoning.
Figure\,\ref{fig:stage_task_guide} summarizes the design overview of these QA categories.
\begin{itemize}[leftmargin=*, itemsep=2pt, topsep=0pt]
    \item \textbf{\textcolor{RoyalBlue}{Dynamic Relativity} QAs}:
    An agent should understand and reason from another agent's viewpoint (reference frame) to communicate spatial information effectively. This requires cross-agent viewpoint transformations to determine a target's direction and distance relative to either agent.
    \item \textbf{\textcolor{DarkOrange}{Regional} QAs}: Common places are often described in terms of semantic regions, such as kitchens and living rooms, which the agent should recognize to interact with other agents. As agents move through these regions, they require jointly reasoning about both regions and dynamics over time.
    \item \textbf{\textcolor{ForestGreen}{Path Reasoning} QAs}: The physical environment, such as walls and furniture, provides spatial context for an agent's movement. The dynamic agent should understand its physical surroundings and reason about actual or hypothetical paths in relation to nearby objects or agents.
\end{itemize}

To evaluate spatial intelligence across these categories, we construct multiple tasks within each category, providing complementary assessments of spatial reasoning involving both the egocentric agent (camera wearer) and the other agent.
We will release \ourbench, including the annotation pipeline and assets for the benchmark generation and evaluation.

\begin{figure}[!t]
    \centering
    \includegraphics[width=\linewidth]{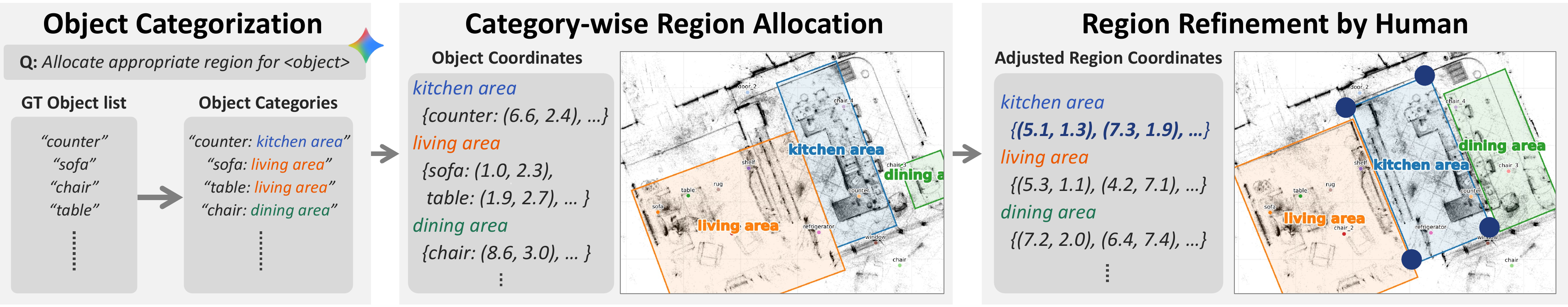}
    \captionsetup{font=footnotesize}
    \vspace{-15pt}
    \caption{\textbf{Region annotation process.} \textit{\textbf{(Left)}} For each scene, we prompt an LLM to assign a region category to every ground-truth object.
    \textbf{\textit{(Middle)}} We group objects by category and use their coordinates to automatically compute the smallest enclosing rectangle for each group, with its sides parallel to the walls.
    \textbf{\textit{(Right)}} Human annotators cross-check the egocentric video and 2D floormap and refine these initial rectangles by adjusting their vertices to obtain region polygons. These are only used for the QA synthesis, not provided to the model.
    }
    \label{fig:region_annotation}
\end{figure}

\subsubsection{Dynamic Relativity QA Synthesis}
\label{subsec:dynamic_relativity_synthesis}

Understanding other agents' viewpoints is essential in everyday interactions. For example, an object to one person's right may lie to another's left when they face each other. To assess this capability, we construct tasks that require models to infer the direction or distance of an imaginary object specified in one agent's reference frame relative to the other agent, requiring a viewpoint transformation (Figure\,\ref{fig:stage_task_guide}\textbf{(a)}).

The task involves two agents: \textit{Ego}, the camera wearer, and \textit{Exo}, the other person in the same scene. We utilize a sound event to specify the query time, as in SAVVY-Bench \cite{chen2025savvy}, and assume an imaginary object at a specific angle and radial distance in one agent's viewpoint. The task asks for this object's direction/distance from the other agent's viewpoint. This is subdivided into \textbf{Ego-to-Exo}, which describes an imaginary object from \textit{Ego}'s perspective and asks about \textit{Exo}'s viewpoint, and \textbf{Exo-to-Ego} in the inverse manner. 
Due to blurred faces in the given dataset, we specify a hypothetical facing direction of \textit{Exo} in each question.

\subsubsection{Regional QA Synthesis}
\label{subsec:regional_synthesis}

Humans often refer to semantic regions, such as living rooms and bedrooms, rather than precise coordinates when communicating spatial information. Therefore, an agent must understand these regional terms and relate them to the positions of agents. We construct region-aware tasks that require joint reasoning about semantic regions and agents' movements over time.
To evaluate these requirements, we first annotate regions by considering the arrangement of key objects and their semantics, utilizing an off-the-shelf model and human refinement (Figure\,\ref{fig:region_annotation}), and then construct region-based tasks with temporal variants, as presented in Figure\,\ref{fig:stage_task_guide}\textbf{(b)}.  \textbf{Location Awareness} asks for the region where an agent is located at the sound event moment, and \textbf{Visit History} asks for the order of visited regions or the unvisited region for a dynamic agent during the time interval between two sound events.

\subsubsection{Path Reasoning QA Synthesis}
\label{subsec:path_reasoning_synthesis}

Moving through real-world environments requires understanding surrounding objects, walls, and other moving agents. We construct \textit{path reasoning} QAs to assess spatial reasoning about agents' movement paths and their relations to the surrounding environment.
We consider two estimation settings (Figure\,\ref{fig:stage_task_guide}\textbf{(c)}), reasoning about physical objects along a hypothetical finite straight path, and estimating the length of an observed agent's trajectory, which is constrained by physical surroundings. \textbf{Line-Path Search} asks for the closest object to the straight-line segment from the camera wearer to the given target. This is also divided into static and dynamic variants, whether the target is a specific object or the other agent, respectively. \textbf{Trajectory Distance} requires a pair of distances traveled by both the camera wearer and the other agent during the sound interval.

\begin{figure}[!t]
    \centering
    \includegraphics[width=\linewidth]{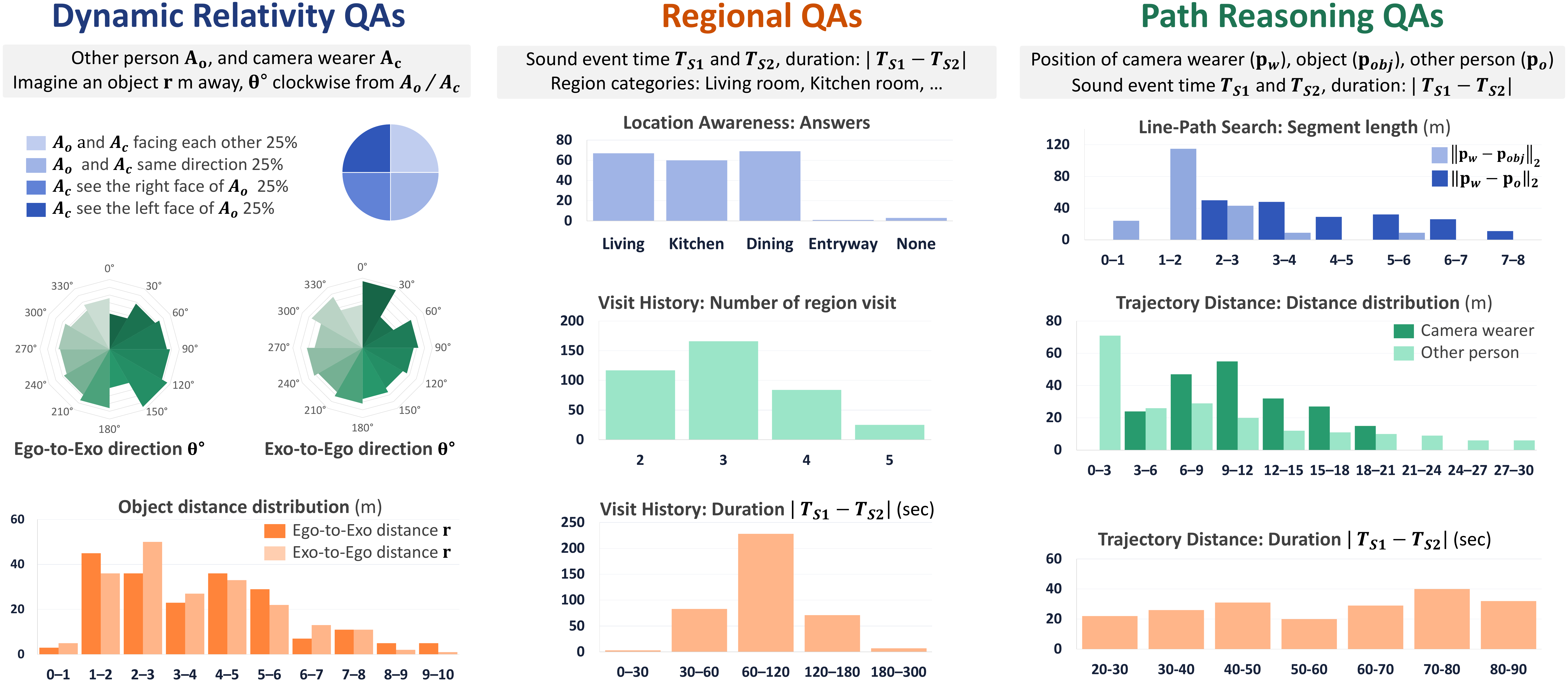}
    \captionsetup{font=footnotesize}
    \vspace{-15pt}
    \caption{\textbf{\ourbench configurations and distributions.} Unlike SAVVY-Bench, whose answers are heavily focused on front-side ($-90^\circ\!\sim\!90^\circ$) and short distances (< 1m), \ourbench is constructed with a broader range of answer cases, covering various spatial configurations and temporal intervals.}
    \label{fig:benchmark_configuration}
\end{figure}

\section{Experiments and Results}
\label{sec:experiments}

\subsection{Setup}

\paragraph{Benchmark and metrics.}
We evaluate \ourmodel on SAVVY-Bench~\cite{chen2025savvy} and our \ourbench, which are both derived from the AEA dataset~\cite{lv2024aria}. SAVVY-Bench comprises over 1,500 QA pairs and \ourbench contains 1,988 QA pairs drawn from 55 indoor daily-life recordings. Specifically, \ourbench comprises 800 \emph{dynamic relativity}, 592 \emph{regional}, and 596 \emph{path reasoning} questions. Statistical details can be found in Figure\,\ref{fig:benchmark_configuration}. All tasks except distance estimation are formulated as multiple-choice problems (MCPs) and evaluated using exact-match accuracy. Distance tasks ask for a numerical value in meters and utilize absolute mean relative accuracy (AbsMRA), with thresholds from 0.1\,m to 1.0\,m.

\paragraph{Baselines and models.}
For evaluation, we compare \ourmodel and the egocentric-video-only strategy with an identical AV-LLM. On SAVVY-Bench, we utilize both open-source and proprietary state-of-the-art models, specifically the Qwen3-Omni model~\cite{xu2025qwen3} and Gemini-2.5-Flash/Pro~\cite{comanici2025gemini}, where the Pro model was also used in the SAVVY framework and its results were reported. \ourbench tests with Gemini-3.6-Flash as the AV-LLM, which is an off-the-shelf model with strengths in video understanding and spatial reasoning, considering the complexity and variety of the contained tasks. All models are evaluated in a zero-shot manner without any fine-tuning.

\subsection{Results on \ourbench}
\label{subsec:saved_results}
\label{subsec:dynamic_relativity}

\ourmodel{} improves all three category averages on \ourbench{} (Table\,\ref{tab:ourbench}), raising the overall score from 58.3\% to 65.3\%. The gains span cross-agent direction, region-aware, and path-related spatial reasoning.

\begin{table}[!t]
    \centering
    \small
    \captionsetup{font=footnotesize}
    \caption{\textbf{\ourbench results with Gemini-3.6-Flash.} \ourmodel improves over the baseline in all categories. Subscripts indicate the difference from Baseline. *Pooled Visit Order and Unvisited questions. $^\dagger$Ground-truth position of the sound source onto the map. $^\ddagger$Partial GT plus the ground-truth object names on the map.}
    \label{tab:ourbench}
    \label{tab:ourbench_oracle}
    \vspace{-5pt}
    \addtolength{\tabcolsep}{1pt}
    \resizebox{.9\textwidth}{!}{%
    \begin{tabular}{l|cc|cc}
        \toprule
        Task & Baseline & \textbf{\ourmodel} & Partial GT Map$^\dagger$ & Full GT Map$^\ddagger$ \\
        \midrule
        \rowcolor{gray!15}
        \multicolumn{5}{l}{\textcolor{RoyalBlue}{\textbf{Dynamic Relativity}}} \\
        \hspace{1em}Ego-to-Exo Direction & 56.0 & \textbf{65.0}\gain{9.0} & 76.5\gain{20.5} & 72.0\gain{16.0} \\
        \hspace{1em}Ego-to-Exo Distance (AbsMRA) & 34.3 & \textbf{34.6}\gain{0.3} & 47.9\gain{13.6} & 45.9\gain{11.6} \\
        \hspace{1em}Exo-to-Ego Direction & 32.0 & \textbf{50.5}\gain{18.5} & 50.5\gain{18.5} & 53.5\gain{21.5} \\
        \hspace{1em}Exo-to-Ego Distance (AbsMRA) & \textbf{46.4} & 44.7\loss{1.7} & 57.5\gain{11.1} & 55.0\gain{8.6} \\
        \hspace{1em}\textit{4-task average} & 42.2 & \textbf{48.7}\gain{6.5} & 58.1\gain{15.9} & 56.6\gain{14.4} \\
        \midrule
        \rowcolor{gray!15}
        \multicolumn{5}{l}{\textcolor{DarkOrange}{\textbf{Regional}}} \\
        \hspace{1em}Location Awareness & 92.0 & \textbf{94.5}\gain{2.5} & 96.0\gain{4.0} & 96.5\gain{4.5} \\
        \hspace{1em}Visit History* & 86.7 & \textbf{92.3}\gain{5.6} & 89.8\gain{3.1} & 97.2\gain{10.5} \\
        \hspace{1em}\textit{2-task average} & 89.4 & \textbf{93.4}\gain{4.0} & 92.9\gain{3.5} & 96.8\gain{7.4} \\
        \midrule
        \rowcolor{gray!15}
        \multicolumn{5}{l}{\textcolor{ForestGreen}{\textbf{Path Reasoning}}} \\
        \hspace{1em}Line-Path Search (static) & 60.2 & \textbf{77.0}\gain{16.8} & 78.1\gain{17.9} & 77.6\gain{17.4} \\
        \hspace{1em}Line-Path Search (dynamic) & 73.0 & \textbf{78.0}\gain{5.0} & 87.5\gain{14.5} & 91.0\gain{18.0} \\
        \hspace{1em}Trajectory Distance (AbsMRA) & 16.9 & \textbf{25.0}\gain{8.1} & 25.0\gain{8.1} & 28.2\gain{11.3} \\
        \hspace{1em}\textit{3-task average} & 50.0 & \textbf{60.0}\gain{10.0} & 63.5\gain{13.5} & 65.6\gain{15.6} \\
        \midrule
        \textbf{Overall (QA-weighted)} & 58.3 & \textbf{65.3}\gain{7.0} & 69.8\gain{11.5} & 71.3\gain{13.0} \\
        \bottomrule
    \end{tabular}
    }
\end{table}

\subsubsection{Dynamic Relativity: Cross-Agent Reasoning}
\label{subsec:dynamic_fov}

\begin{wrapfigure}[15]{R}{0.35\textwidth}
    \vspace{-12pt}
    \centering
    \includegraphics[width=\linewidth]{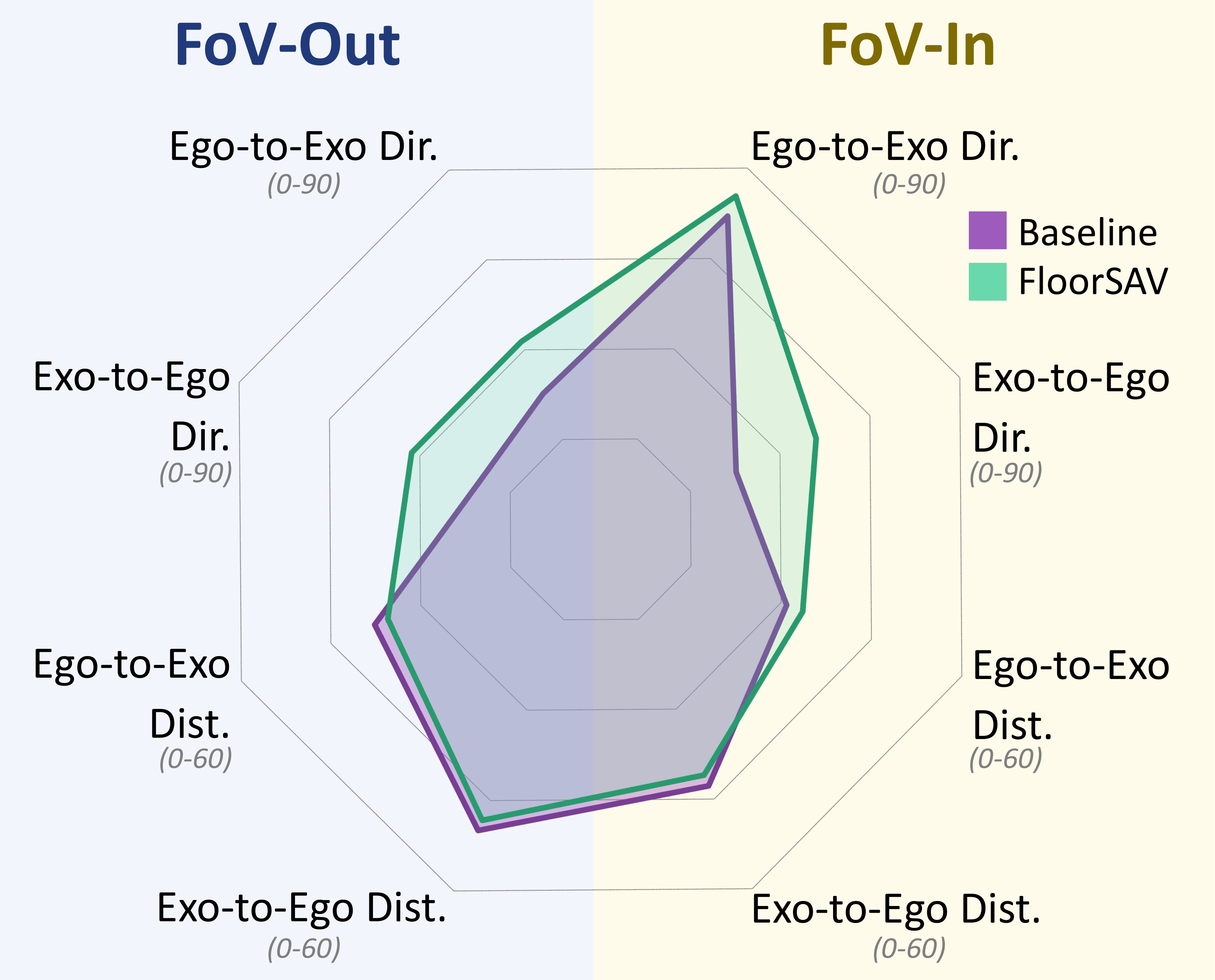}
    \captionsetup{font=footnotesize}
    \vspace{-16pt}
    \caption{\textbf{Dynamic relativity by visibility.} \ourmodel helps more on FoV-out (\ie other agent invisible at the queried moment). Both strategies are weaker on \textit{Exo-to-Ego} direction tasks.}
    \label{fig:dynamic_fov}
\end{wrapfigure}

We analyze \textit{dynamic relativity} QAs using the benchmark's FoV-in/out annotations, which indicate the other agent's visibility at the sound-event query time (Figure\,\ref{fig:dynamic_fov}).

\paragraph{Relative direction reasoning.}
\ourmodel improves all direction tasks under all visibility conditions. The overall direction gain is larger in FoV-out (+15.0 vs. +12.5) where the other agent does not exist in the egocentric view at the querying moment. While \ourmodel improves the baseline, we find that both strategies show lower performance on \textit{Exo-to-Ego} than on \textit{Ego-to-Exo}, although both tasks require the same information: the two agents' poses and headings. This suggests that the direction of cross-agent viewpoint transformation may affect the difficulty of spatial reasoning for AV-LLMs.

\paragraph{Metric distance reasoning.}
Results on the distances are less consistent, since they are directly affected by errors in the sound-based estimation of the other agent's position. Improvements with accurate other-person position (Section~\ref{subsec:gtmap_ablation}) support localization noise as a main bottleneck for distance tasks.

\subsubsection{Regional: Bridge Semantic Region with Dynamic Agents}

Both the baseline and \ourmodel demonstrate fair performance on \emph{regional} QAs, where dialogue may provide clues to infer the current region at an audio timestamp (Table\,\ref{tab:regional_audio_only}).
Nevertheless, \ourmodel improves \textit{Location Awareness} to 94.5\% and the regional average from 89.4\% to 93.4\%. The improvement is more pronounced on \textit{Visit History}, reaching 92.3\% (+5.6) across ordered and unvisited tasks, which require retaining and ordering visits over the temporal axis. This result supports the effectiveness of the floormap for integrating agents' positions, key objects, and the map as an explicit representation.

\subsubsection{Path Reasoning: Spatial Awareness on Physical Environment}

\ourmodel demonstrates gains on all \textit{path reasoning} tasks as shown in Table\,\ref{tab:ourbench}. Specifically, improvements on both \textit{Line-Path Search} tasks support the advantages of rendered key objects as a consequence of semantic grounding. \textit{Trajectory Distance} improves from 16.9 to 25.0 AbsMRA; however, sparse and inaccurate sound localization magnifies the discrepancy between the derived and true distances. Moreover, historical path rendering improves the \textit{Trajectory Distance} performance (Table\,\ref{tab:historical_path_ablation}), showing the benefit of explicit motion history.

\subsection{Results on SAVVY-Bench}
\label{subsec:main_results}

\paragraph{\ourmodel improves performance across AV-LLMs.}
Table\,\ref{tab:savvy_benchmark} shows that \ourmodel improves overall scores on SAVVY-Bench for all AV-LLMs, Qwen3-Omni-30B, Gemini-2.5-Flash, and Pro with semantic grounding. Especially large gains are shown in the \textit{exocentric direction}, a noted challenge for AV-LLMs~\cite{chen2025savvy}.
All baselines fall below the random-chance level (32.1\%), whereas \ourmodel achieves 52.8\% with Gemini-2.5-Pro.
While slightly degraded on distance tasks, studies with ground-truth agents' positions and object list (Section~\ref{subsec:gtmap_ablation}) suggest inaccurate acoustic estimates and unaligned object labels may degrade performance.

Performance is also inherently limited by AV-LLMs' temporal localization of target sound events, biases in relative-position reasoning that ignore video cues, and lack of cognitive capability to fully interpret geometric 2D floormaps. Appendix~\ref{app:qualitative} and Appendix~\ref{app:limitations} detail these failure cases and limitations, respectively.

\begin{table}[!t]
    \centering
    \begin{minipage}[t]{0.6\textwidth}
        \centering
        \small
        \captionsetup{font=footnotesize}
        \caption{Audio-only \textit{regional} QA accuracy (\%) suggests that the audio contains regional context that the AV-LLM can utilize.}
        \label{tab:regional_audio_only}
        \vspace{-5pt}
        \renewcommand{\arraystretch}{0.98}
        \addtolength{\tabcolsep}{-1pt}
        \resizebox{\linewidth}{!}{%
        \begin{tabular}{l|cccc}
            \toprule
            Input & Random & \!\!Audio-only\!\! & \!Baseline\! & \textbf{\ourmodel} \\
            \midrule
            Location Awareness & 25.0 & 72.0 & 92.0 & 94.5 \\
            Visit History      & 25.0 & 62.2 & 86.7 & 92.3 \\
            \bottomrule
        \end{tabular}%
        }
    \end{minipage}
    \hfill
    \begin{minipage}[t]{0.375\textwidth}
        \centering
        \small
        \captionsetup{font=footnotesize}
        \caption{Historical camera wearer path ablation on Trajectory Distance (AbsMRA).}
        \label{tab:historical_path_ablation}
        \vspace{-5pt}
        \renewcommand{\arraystretch}{0.95}
        \resizebox{\linewidth}{!}{%
        \begin{tabular}{l|c}
            \toprule
            Historical path & AbsMRA $\uparrow$ \\
            \midrule
            Without & 22.0 \\
            With (about 300 frames) & \textbf{25.0} \\
            \bottomrule
        \end{tabular}
        }
    \end{minipage}
\end{table}

\begin{table}[!t]
    \centering
    \small
    \captionsetup{font=footnotesize}
    \caption{\textbf{SAVVY-Bench results.} For \ourmodel, Gemini also supplies semantic grounding. SAVVY scores are from \citet{chen2025savvy}.
    Direction uses accuracy, and distance uses AbsMRA. Avg.\ is the four-task mean.}
    \label{tab:savvy_benchmark}
    \vspace{-5pt}
    \renewcommand{\arraystretch}{0.95}
    \resizebox{\textwidth}{!}{%
    \begin{tabular}{l|cc|cc|ccc|cc}
        \toprule
        & \multicolumn{2}{c|}{Qwen3-Omni-30B} & \multicolumn{2}{c|}{Gemini-2.5-Flash} & \multicolumn{5}{c}{Gemini-2.5-Pro} \\
        \cmidrule(lr){2-3}\cmidrule(lr){4-5}\cmidrule(lr){6-10}
        Task & Base & \textbf{\ourmodel} & Base & \textbf{\ourmodel} & Base & SAVVY & \textbf{\ourmodel} & Partial$^\dagger$ & Full$^\ddagger$ \\
        \midrule
        Ego Direction & 71.7 & 72.8 & 74.2 & 71.3 & 75.2 & 84.7 & 75.8 & 70.2 & 68.3 \\
        Ego Distance & 59.4 & 61.8 & 49.7 & 55.4 & 59.6 & 62.9 & 55.4 & 59.0 & 61.0 \\
        Exo Direction & 29.4 & 32.9 & 29.8 & 44.5 & 31.7 & 44.0 & 52.8 & 52.3 & 69.9 \\
        Exo Distance & 31.8 & 31.0 & 29.0 & 32.2 & 37.0 & 40.2 & 34.9 & 37.9 & 52.2 \\
        \midrule
        \textbf{Avg.} & 48.1 & 49.6 & 45.7 & 50.9 & 50.9 & 58.0 & 54.7 & 54.9 & 62.8 \\
        \bottomrule
    \end{tabular}%
    }
\end{table}

\subsection{Studies with Ground-Truth 2D Floormap}
\label{subsec:gtmap_ablation}

Although we demonstrate the benefits of \ourmodel across the spatial benchmarks, errors in acoustic estimates and object mapping may restrict the potential of \ourmodel. Therefore, we conduct studies with two versions of GT 2D floormap, \textit{\textbf{partial GT map}} and \textit{\textbf{full GT map}}. Both floormaps replace acoustic estimates with the GT position of the other agent, while \textit{full GT map} additionally renders ground-truth objects. These studies show the potential of \ourmodel with accurate components.

\paragraph{Ground-truth position of the other agent.}
Both \textit{partial GT map} and \textit{full GT map} considerably improve distance-based tasks (Tables \,\ref{tab:ourbench} and \ref{tab:savvy_benchmark}), \textit{Line-Path Search (dynamic)}, and \textit{path reasoning} overall (Table\,\ref{tab:ourbench}). Improvement of these agent-position-based tasks implies the potential of \ourmodel with accurate agent positions.

\paragraph{Ground-truth object list.}
\textit{Full GT map} rendered with GT objects improves performance on all \textit{regional} QAs (Table \,\ref{tab:ourbench}) which inherently depend on accurate region recognition. This suggests that when accurate object labels are provided, \ourmodel effectively associates their spatial locations with semantic regions.

\section{Conclusion}

We introduce \textbf{\ourmodel}, a task-agnostic, training-free framework that supports spatial reasoning in AV-LLMs through synchronized 2D floormaps integrating camera tracking, spatial audio, and object grounding. We also introduce \textbf{\ourbench}, an egocentric spatial audio-visual benchmark with dynamic agents, spanning complementary capabilities essential to real-world agent interactions. Evaluations on SAVVY-Bench and \ourbench demonstrate the benefits of \ourmodel across single-viewpoint reasoning, cross-agent viewpoint transformation, regional reasoning over agents' movements, and path-aware reasoning grounded in physical surroundings. Studies on GT floormaps further demonstrate the framework's potential with accurate data. The benefits of floormaps may extend beyond the training-free regime to incorporate them into model training, enabling models to internalize them and potentially achieve more general and capable spatial intelligence.

\bibliographystyle{plainnat}
\bibliography{references}

\clearpage
\appendix

\section{Implementation Details}
\label{app:implementation}

\subsection{Floormap Rendering Configuration}
\label{app:floormap_rendering}

\paragraph{Base map construction.}
To construct the global base layout, we process the raw 3D point clouds provided by the AEA dataset~\cite{lv2024aria} and apply a sequence of spatial filters to remove structural outliers.
(1) First, we bound the vertical height to the 10th--99th percentile with a 5\,cm inset margin to discard floor and ceiling artifacts. The XY bounds span the 1st--99th percentiles expanded outward by 0.5\,m. To ensure geometric reliability, we retain only points with reconstruction uncertainty \texttt{dist\_std} below 0.02 when this field is available.
(2) Some regions contain dense point accumulations, for example where a person remains in place during recording. To reduce visual clutter, we divide the XY plane into 0.05\,m grid cells and identify cells containing at least 500 points as potential dynamic accumulations. These dense regions are rendered as a Gaussian-blurred gray heatmap over a sparse black scatter plot of the filtered point cloud, which represents the room structure and object boundaries.
(3) Each floormap frame is generated at a high resolution on a white background to preserve fine geometric details when injected into the model. To provide an absolute spatial reference, the coordinate axes are configured with metric units, featuring a 1\,m grid and integer tick labels anchored to the scene's minimum bounds.

\paragraph{Camera state.}
To explicitly visualize the observer's spatial state, we render the current camera position as a distinct red dot, accompanied by a prominent red directional arrow indicating the heading. We further approximate the camera's visual coverage by projecting a yellow wedge that spans a $\pm 50^\circ$ FoV.
For temporal navigation context, the observer's historical motion is smoothed and plotted as a blue continuous polyline representing the most recent 300 camera frames (approximately 15 seconds).

\paragraph{Acoustic tracking and spatial cues.}
To integrate dynamic sound sources, we follow the acoustic tracking pipeline from SAVVY~\cite{chen2025savvy}, where each sound source position is defined by an egocentric angle estimated with SRP-PHAT~\cite{dibiase2000high} and a distance estimated from the coherent-to-diffuse ratio (CDR)~\cite{schwarz2015coherent}. Spatial localization uses four Aria microphone channels among seven~\cite{chen2025savvy} at 48\,kHz before monaural conversion for AV-LLM input. Non-overlapping 0.25\,s chunks yield four observations per second, timestamped at midpoints of each interval. These estimates serve as explicit spatial context for the AV-LLM rather than directly determining the final answer.

\noindent\textbf{SRP-PHAT direction estimation.} For microphone signals with Fourier transforms $X_m(f)$ and $X_n(f)$, the phase-normalized cross-correlation is
\begin{equation}
    R_{mn}[\ell]=\operatorname{Re}\!\left\{
    \mathcal{F}^{-1}\!\left[
    \frac{X_m(f)X_n(f)^*}{|X_m(f)X_n(f)^*|+\epsilon}
    \right][\ell]\right\},
    \label{eq:appendix_gcc_phat}
\end{equation}
where $\epsilon$ prevents division by zero. Given microphone coordinates $\mathbf{m}_m,\mathbf{m}_n$ and candidate direction $\mathbf{u}(\phi)$, the expected delay is $\tau_{mn}(\phi)=(\mathbf{m}_n-\mathbf{m}_m)^\top\mathbf{u}(\phi)/c$, with $c=343$\,m/s. SRP-PHAT selects the azimuth maximizing the summed pairwise response:
\begin{equation}
    \hat\phi=\arg\max_{\phi}\sum_{m<n}
    R_{mn}\!\left[\operatorname{round}\bigl(f_s\tau_{mn}(\phi)\bigr)\right].
    \label{eq:appendix_srp_phat}
\end{equation}
Here $f_s$ is the sampling rate. The implementation searches $-180^\circ$ to $180^\circ$ at $1^\circ$ increments and restricts cross-correlation lags to $\pm1$\,ms.

\noindent\textbf{CDR-based distance estimation.} Source distance is estimated from the approximate acoustic relation $D_t^2\,\mathrm{CDR}_t\approx K$. For visual distance estimates $D_t$ at calibration times retained after outlier filtering, $t\in\mathcal{I}$, SAVVY~\cite{chen2025savvy} provides the calibration and distance conversion as
\begin{equation}
    \hat K=\arg\min_K\sum_{t\in\mathcal{I}}
    \bigl(D_t^2\,\mathrm{CDR}_t-K\bigr)^2,
    \qquad \hat d_t=\sqrt{\frac{\hat K}{\mathrm{CDR}_t}}.
    \label{eq:appendix_cdr_distance}
\end{equation}
For each recording, we take the median of the positive calibration constants (K) pre-calculated by SAVVY, which are utilized to convert CDR values into sound-source distances for all downstream tasks.

\noindent\textbf{Projection and sound-marker aggregation.} Each observation is aligned with its nearest camera pose. Let $\mathbf{c}_t$ be the camera's XY position, $\mathbf{f}_t$ its normalized XY forward vector, and $\mathbf{r}_t=(f_{t,y},-f_{t,x})^\top$ its rightward vector. The exported camera-relative angle $\alpha_t$, positive clockwise, gives the global coordinates estimate
\begin{equation}
    \hat{\mathbf{s}}_t=\mathbf{c}_t+\hat d_t
    \bigl(\cos\alpha_t\,\mathbf{f}_t+\sin\alpha_t\,\mathbf{r}_t\bigr).
    \label{eq:appendix_audio_projection}
\end{equation}
We discard predictions within the extreme frontal ($|\alpha_t|\leq8^\circ$) or rear ($|\alpha_t|\geq172^\circ$) ranges. We calculate a circular mean of global frame bearings and an arithmetic mean of distances within $\pm1$\,s of the output frame's timestamp. The green dot is placed at the resulting bearing and radial distance from the rendered camera position on the floormap. The marker is omitted when no valid observation is available.

To construct the synchronized video input, we uniformly sample $N=128$ frames across the entire duration of both the egocentric RGB footage and the 2D floormap sequence. Each frame corresponds to an equal-duration temporal bin, with the camera's position and heading taken from the nearest pose to its midpoint. The representative sound marker is obtained by averaging acoustic estimates within a $\pm1$\,s window around the bin midpoint rather than over the full bin.

\paragraph{Semantic grounding via object mapping.} To bridge the AV-LLM's semantic and spatial awareness, we implement a multi-stage pipeline that merges semantic labels into the floormap, as described in Figure\,\ref{fig:semantic_grounding}.

\begin{enumerate}[leftmargin=*, itemsep=2pt, topsep=2pt, label=(\arabic*)]
    \item \textbf{Frame selection:} For each scene in the AEA dataset, we select a total of 30 keyframes through three rounds of greedy maximum-coverage selection, with 10 frames selected per round from video sampled at 1 FPS. At each selection step within a round, we choose the frame that maximizes the newly covered floormap area under a circular-sector FoV defined by $-50^\circ \le \theta \le 50^\circ$ and $0 \le r \le 5$ meters. The covered area is reinitialized at the beginning of each round, whereas frames selected in previous rounds are excluded from the candidate set. This procedure yields three distinct sets of 10 frames.
    \item \textbf{Object visual mapping:} We leverage the VLM's visual grounding capabilities to detect representative objects and project them onto the 2D floormap. For each selected frame, we provide dual image inputs: the egocentric view and the corresponding floormap view, which is rotated so that the camera heading aligns with the upward (+$y$) direction, simplifying directional alignment during coordinate mapping. We design the prompt as presented in Figure\,\ref{fig:appendix_semantic_grounding_prompt} to utilize the VLM's spatial priors and to retain multiple observations for the subsequent \textit{\textbf{object filtering}} step, including repeated detections of the same instance across frames and distinct instances that share a semantic label.
    \item \textbf{Object filtering:} Because VLM detections may contain inconsistent names and geometrically implausible coordinates, we consolidate the observations in three stages. \textit{(i)} We first normalize class names by lowercasing and standardizing separators, plurals, and common synonyms (\eg \textit{television} to \textit{tv} and \textit{couch} to \textit{sofa}). We then transform each prediction from its heading-aligned local floormap coordinates into the global coordinate system. Predictions outside the point-cloud bounds with a 0.5-meter buffer are discarded, as are predictions supported by fewer than 100 point-cloud points within a 0.5-meter radius. \textit{(ii)} For each normalized class, we construct a graph that connects detections separated by at most 2 meters and obtain spatial groups as its connected components. This procedure merges repeated observations of the same object while preserving spatially separated instances that share a class name. \textit{(iii)} We retain only components supported by at least three detections and use the mean of their global coordinates as the final object location.
    \item \textbf{Floormap Semantic Grounding:} Consequently, the consolidated instances and their corresponding semantic labels are mapped onto the final floormap, with spatially distinct instances of the same class retained separately. This semantically grounded floormap enhances the AV-LLM's spatial reasoning performance within our \ourmodel framework.
\end{enumerate}

\begin{figure}[!t]
    \centering
    \footnotesize
    \begin{tcolorbox}[width=\linewidth, sharp corners=all, colback=gray!10, boxrule=0.3mm, top=1mm, bottom=1mm, left=1mm, right=1mm]
    \ttfamily\setlength{\parskip}{0.25ex}
    {\textbf{image 1:  \# Egocentric}} \\
    {<|vision\_token|>} \\[1ex]
    {\textbf{image 2:  \# Floormap}} \\
    {<|vision\_token|>} \\[1ex]

    {\textbf{Task Instructions:}} \\
    {1. Estimate the direction and distance of all objects in the egocentric image.} \\
    {2. Utilizing the results from Step 1 alongside the objects visible in the floormap, estimate the exact position of all objects on the floormap.} \\[1ex]

    {\textbf{Context:}} \\
    \hspace*{0.5em}{- \textbf{Red Dot \& Arrow:}} Expresses the position and heading direction of the egocentric view; the length of the red arrow is strictly 1 meter. \\
    \hspace*{0.5em}{- \textbf{Yellow Area:}} Represents the camera field of view (FoV) of the egocentric view (egocentric angle range: -50 to 50 degrees, maximum distance: 5 meters). \\
    \hspace*{0.5em}{- \textbf{Crucial Step:}} Matching the environment context of the egocentric view to the shapes of objects and the overall structure of the floormap within the FoV is a key phase; do not ignore it. \\
    \hspace*{0.5em}{- \textbf{Object Density:}} Only consider representative and high-confidence estimations (at most 5 objects). Focus heavily on major structures such as furniture, doors, and windows, excluding minor or trivial items. \\
    \hspace*{0.5em}{- \textbf{Naming Convention:}} Name objects using generic, concise nouns; avoid specific brand/proper names and excessively long descriptions. \\[1ex]

    {\textbf{Output Formatting (STRICTLY FOLLOW):}} \\
    {Print out only a list of JSON objects formatted as follows:} \\
    {\noindent\texttt{[}} \\
    \hspace*{1.0em}{\texttt{\{"object", direction", "distance", "coordinates"\}\}, \dots}} \\
    {\noindent\texttt{]}}
    \end{tcolorbox}
    \vspace{-10pt}
    \caption{\textbf{Pseudo prompt for object visual mapping stage.} The prompt is designed to leverage the VLM's visual grounding prior on egocentric view images, enabling accurate coordinate mapping, the extraction of representative visual landmarks, and the simplification of semantic labels.}
    \label{fig:appendix_semantic_grounding_prompt}
\end{figure}

\begin{figure}[t]
    \centering
    \includegraphics[width=\linewidth]{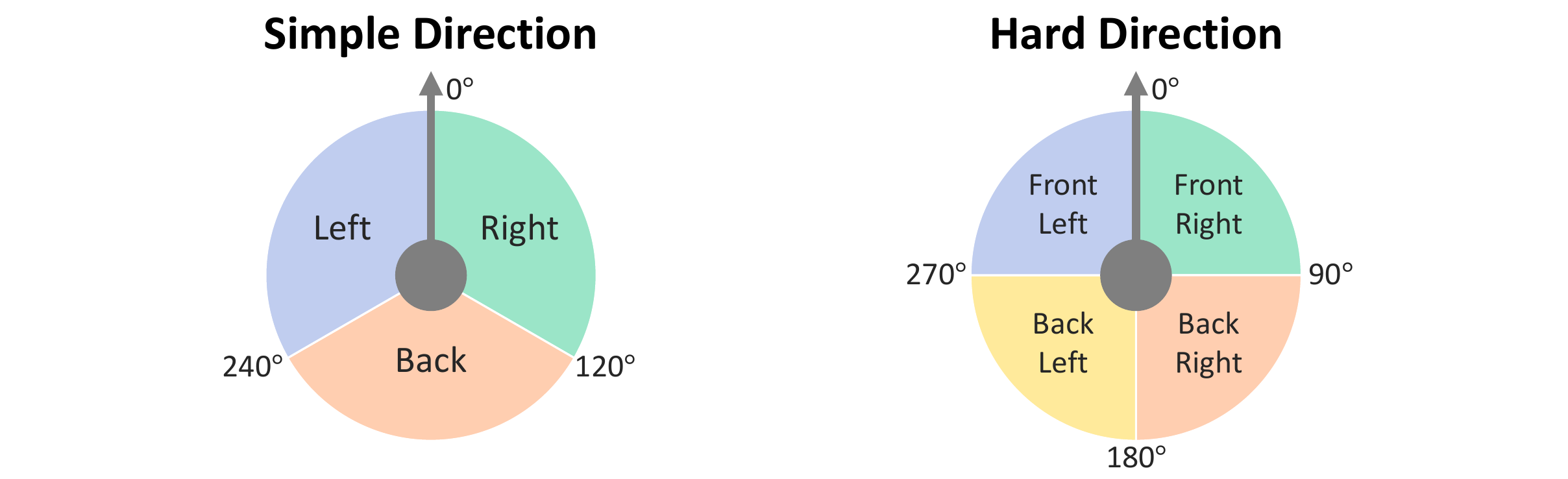}
    \vspace{-15pt}
    \caption{\textbf{Simple and hard direction configurations.} Relative to the target agent, simple direction utilizes three options (left, right, back), and hard direction utilizes four (front-left, front-right, back-left, back-right).}
    \label{fig:appendix_direction_configurations}
\end{figure}

\subsection{Evaluation Details}
\label{app:evaluation_details}

During inference, we employ deterministic greedy decoding with a temperature of 0.0 to ensure consistent evaluations. \ourmodel involves no training of any kind. The proprietary Gemini models are queried through their public API, and the open-source model, Qwen3-Omni-30B, is run for inference only on four NVIDIA RTX A6000 GPUs.
To accommodate our dual-stream \ourmodel framework (see Appendix~\ref{app:prompting_strategies}), the input prompt for the Qwen-Omni models is explicitly structured to integrate multiple synchronized modalities. Specifically, we sequentially construct the prompt by feeding the egocentric RGB footage and the rendered 2D floormap as two independent video inputs, following the extracted monaural audio track, as illustrated in Figure~\ref{fig:sample_prompt}. For details, please refer to the full instruction prompt provided in Appendix~\ref{app:prompts}. These multiple streams are strictly synchronized, ensuring the model can temporally align the first-person visual appearance directly with the dynamic geometric layout of the top-down map as well as the audio track.

For comparison with prior work, we utilize the QA configuration of SAVVY-Bench~\cite{chen2025savvy}, which encompasses four distinct task types: \textbf{egocentric direction}, \textbf{egocentric distance}, \textbf{exocentric direction}, and \textbf{exocentric distance}.
Egocentric tasks evaluate spatial relations directly from the camera wearer's first-person perspective, whereas \textit{exocentric} tasks require the model to deduce relationships from a hypothetical third-person viewpoint anchored to specific static objects within the scene.
\begin{itemize}[leftmargin=*, itemsep=0pt]
    \item \textbf{Egocentric QA:} The spatial reference frame is anchored to the camera wearer. The origin is the camera's current position, and the forward axis aligns with the camera's heading direction. The model must localize the sounding target relative to the observer's perspective.

    \item \textbf{Exocentric QA:} The spatial reference frame is anchored to static objects in the environment. A hypothetical observer (\ie robot) is positioned at a specific \textit{reference object} (\eg fireplace) and faces a \textit{facing object} (\eg TV). This establishes a new third-person coordinate system where the model must mentally transform its perspective to estimate the target's location.
\end{itemize}

Directional tasks are posed as multiple-choice questions with either three options (\ie left, right, back) for simple layouts or four options (\ie front-left, front-right, back-left, back-right) for harder layouts (Figure\,\ref{fig:appendix_direction_configurations}), whereas distance tasks require a direct numerical estimate in meters.

In the original benchmark design, the question prompts are heavily supplemented with explicit textual spatial rules, such as defining \textit{120-degree turning sectors} for simple choices or detailing \textit{Cartesian quadrants and XY-axes} for hard choices. However, we empirically found that providing these verbose mathematical definitions often confuses the AV-LLMs, as the models struggle to mentally align the strict text-based geometric rules with the provided visual cues. Therefore, we deliberately strip away or simplify these textual coordinate instructions from the base questions for both the egocentric baseline and \ourmodel. This intentional simplification encourages the model itself to intuitively deduce spatial relationships and orientations directly from the visual geometry of the provided video inputs without being distracted by redundant textual constraints.

\subsection{\ourbench Configuration}
\label{app:ourbench_configuration}

\paragraph{Benchmark overview.}
\ourbench evaluates broad axes of spatial intelligence required for an egocentric agent in real-world scenarios. The benchmark comprises 1,988 QAs across three complementary categories: \textit{Dynamic Relativity} (800), \textit{Regional} (592), and \textit{Path Reasoning} (596), with each category consisting of multiple tasks. These tasks evaluate the spatial capabilities required by each category: \textit{Dynamic Relativity} evaluates cross-agent viewpoint transformations, \textit{Regional} evaluates joint reasoning about regions and dynamic agents, and \textit{Path Reasoning} evaluates the spatial awareness of dynamic agents in physical environments. Queries are grounded at a sound-event moment or within the temporal interval between two distinct sound events and ask task-specific questions about the audio-visual scene with dynamic agents.

\paragraph{Task grouping and metric details.}
\textit{Visit History} pools \textit{Visit Order} and \textit{Unvisited} questions together, with 194 and 198 QAs, respectively. Direction tasks in \textit{dynamic relativity} QAs, \textit{regional} QAs, and \textit{Line-Path Search} tasks use multiple-choice answers evaluated by exact-match accuracy as binary scores, i.e., zero or one. Distance-wise tasks ask for numerical values in meters and are evaluated using AbsMRA, as defined below. Category scores are averages with equal weight per task, with \textit{Visit History} treated as one pooled task. The overall score is weighted by the number of QAs in each task.

\paragraph{Distance metrics: AbsMRA and RelMRA.}
AbsMRA averages accuracies over ten absolute-error thresholds, $\mathcal{T}_{\mathrm{abs}}=\{0.1,0.2,\ldots,1.0\}$\,m, giving partial credit to predictions that satisfy more thresholds. RelMRA instead uses ten relative-error thresholds, $\mathcal{T}_{\mathrm{rel}}=\{0.05,0.10,\ldots,0.50\}$, measuring error relative to the ground-truth distance. For $N$ distance predictions $\hat d_i$ and ground-truth values $d_i$, let $e_i=|\hat d_i-d_i|$. The percentage scores are
\begin{equation}
    \begin{aligned}
        \mathrm{AbsMRA} &= \frac{100}{10N}\sum_{i=1}^{N}
        \sum_{\tau\in\mathcal{T}_{\mathrm{abs}}}\mathbf{1}[e_i\leq\tau], \\
        \mathrm{RelMRA} &= \frac{100}{10N}\sum_{i=1}^{N}
        \sum_{\rho\in\mathcal{T}_{\mathrm{rel}}}
        \mathbf{1}\!\left[\frac{e_i}{\max(d_i,\epsilon)}\leq\rho\right],
    \end{aligned}
    \label{eq:ourbench_distance_metrics}
\end{equation}
where $\mathbf{1}[\cdot]$ is the indicator function and $\epsilon=10^{-9}$ prevents division by zero. Numerical tolerances are applied at threshold boundaries in the implementation. AbsMRA imposes the same sensitivity regardless of the target distance, whereas RelMRA allows proportionally larger absolute errors for longer target distances. For example, a 0.5\,m error corresponds to a 50\% relative error at 1\,m but only 5\% at 10\,m. We use AbsMRA as the primary distance metric in the reported tables; RelMRA provides a complementary measure.

\paragraph{Exception cases.}
Although we strongly restrict the output format, some responses disobey the instruction and answer without following the output format. For those cases, we attempt to recover an answer from the full response. However, responses in which the model fails to provide any answer to the question are marked as \textit{unparsed} and receive zero credit. The following subsections describe the construction and ground-truth definitions for each category.

\paragraph{Temporal alignment and evaluation.}
The benchmark contains 1,988 QAs with temporally aligned other-agent GT. The other agent's trajectory is queried at camera wearer time with the offset between guest and wearer start times from the speech metadata. All evaluated settings use the same QA IDs. Studies with partial/full GT maps utilize synchronized 6-DoF data for both agents.

\subsection{Dynamic Relativity QAs Details}
\label{app:dynamic_relativity}

\paragraph{Task construction.}
We construct the \textit{dynamic relativity} QAs from 50 AEA scenes, using a sound event to identify the query moment in each egocentric video, following the SAVVY-Bench schema. Each question describes an imaginary (hypothetical) object by an angle and a radial distance from one of two moving agents: the camera wearer or the other person.

\paragraph{Object placement and reference frames.}
Task formulation consists of transformation between agents' viewpoints. Let $\mathbf{p}_a$ and $\mathbf{f}_a$ be agent $a$'s global position and unit forward vector at the query time, for $a\in\{w,o\}$, where $w$ denotes the camera wearer (\textit{Ego}) and $o$ the other person (\textit{Exo}). Let $R_{\mathrm{cw}}(\theta)$ denote a clockwise planar rotation by $\theta$. Given a source agent $s$, a target agent $t\neq s$, and a question-specified angular offset $\theta$ and radial distance $r$, the imaginary object's global position is
\begin{equation}
    \mathbf{p}_{obj} = \mathbf{p}_s + r\,R_{\mathrm{cw}}(\theta)\,\mathbf{f}_s ,
    \label{eq:appendix_dr_placement}
\end{equation}
The question asks for the direction or distance of $\mathbf{p}_{obj}$ in the target frame defined by $\mathbf{p}_t$ and $\mathbf{f}_t$. Agent positions and the camera wearer's heading are obtained from the synchronized 6-DoF data in global coordinates. Because faces are blurred in AEA, we explicitly describe the hypothetical facing direction of the other agent $\mathbf{f}_o$ in the question.

\begin{itemize}[leftmargin=*, itemsep=0pt]
    \item \textbf{Ego-to-Exo.} We set $(s,t)=(w,o)$. The object is described by $\theta$ and $r$ relative to the camera wearer's position and heading. The model must transform this description into the other person's reference frame and infer the object's direction relative to the described hypothetical facing direction in question, or its distance from the other person. Here, $\mathbf{f}_o$ defines the target frame for relative direction reasoning.

    \item \textbf{Exo-to-Ego.} We set $(s,t)=(o,w)$. The object is described by $\theta$ and $r$ relative to the other person's position and hypothetical facing direction. The model must transform this spatial information into the camera wearer's reference frame and infer the object's direction relative to the camera heading, or its distance from the camera wearer. Here, $\mathbf{f}_o$ defines the source frame used to place the imaginary object.
\end{itemize}

\paragraph{Ground-truth answers.}
For both transformations, we construct multiple-choice answers, specifically four quadrants (front-left, front-right, back-left, back-right) relative to the target agent's heading direction (Figure\,\ref{fig:appendix_direction_configurations}), whereas the distance is derived from the Euclidean distance $\|\mathbf{p}_{obj}-\mathbf{p}_t\|_2$ in meters. Ground-truth positions are computed from agents' trajectory data provided in the AEA dataset. AV-LLMs receive the question and the egocentric video with its audio track for the baseline, and additionally the rendered 2D floormap video and its guide prompts for \ourmodel.

\paragraph{Dataset statistics.}
The \textit{dynamic relativity} set contains 800 QAs, with 200 per task and 100 per FoV-in/out subset according to the benchmark annotations (Figure\,\ref{fig:dynamic_fov}). FoV-in/out annotations describe the other agent's visibility at the sound-event query time. The overall \textit{dynamic relativity} score is the average of these four task scores (Table\,\ref{tab:ourbench}).

\subsection{Regional QAs Details}
\label{app:regional_details}

\paragraph{Task construction.}
\textit{Regional} QAs require joint spatial reasoning about dynamic agents and semantic regions. The model should recognize regions and align them with agents' positions. We construct tasks along the temporal axis: the moment of a queried sound event or a time interval between two distinct sound events, as \textbf{Location Awareness} and \textbf{Visit History}, respectively. \textit{Location Awareness} is balanced between the camera wearer (\textit{Ego}) and the other agent (\textit{Exo}), with 100 QAs each. \textit{Visit History} contains 200 camera-wearer and 192 other-agent QAs.

\begin{itemize}[leftmargin=*, itemsep=0pt]
    \item \textbf{Location Awareness.} The answer is the name of the region where the target agent, \textit{Ego} or \textit{Exo}, is at the sound event moment described in the question.

    \item \textbf{Visit History.} The question asks about the order of regions visited by the target agent or which regions remain unvisited within the temporal interval between two queried sound events.
\end{itemize}

\paragraph{Region GT annotation pipeline.}
The AEA dataset does not provide region-level semantic annotation. Therefore, we introduce a region GT annotation pipeline that combines an off-the-shelf model and human annotators (Figure\,\ref{fig:region_annotation}).

\begin{itemize}[leftmargin=*, itemsep=0pt]
    \item \textbf{Step 1: Object-wise region categorization.} For each scene, we obtain the ground-truth object labels and positions, then provide the object labels to Gemini 3.6 Flash, asking for a single region category for each object in a strict output format that outputs the region label only. We do not provide a list of regions, in an open-vocabulary manner, to preserve flexibility and scalability.

    \item \textbf{Step 2: Primitive region boundaries.} For each group of objects assigned to the identical region, we draw the smallest rectangle enclosing all object positions, with its edges constrained to be parallel to the surrounding walls in the scene.

    \item \textbf{Step 3: Human refinement.} We render all region rectangles on the 2D floormap, and human annotators check both the floormap and the egocentric video to refine the rectangles into polygons so that the region annotations align more accurately with the physical environment of the scene. All finalized polygon edges are constrained to be parallel to the scene walls.
\end{itemize}

The resulting regions serve as ground-truth for \textit{regional} QAs construction. Region boundaries and category labels are not rendered on the maps provided to \ourmodel, including in the GT map studies. This setting prevents a visual-grounding shortcut in which the model could directly match an agent's position to a rendered region label instead of reasoning about the region from the available scene context, including key objects.

\paragraph{Dataset statistics.}
The evaluated \textit{regional} category contains 592 QAs: 392 \textit{Visit History} questions (194 asking about visit order and 198 asking which regions remain unvisited between two sound events) and 200 \textit{Location Awareness} questions. All \textit{regional} questions use multiple-choice exact-match accuracy. The \textit{Regional} score averages the \textit{Visit History} and \textit{Location Awareness} task scores (Table\,\ref{tab:ourbench}).

\subsection{Path reasoning QAs Details}
\label{app:path_reasoning_details}

\paragraph{Task construction.}
\textit{Path reasoning} QAs evaluate whether the model is aware of its physical surroundings during spatial reasoning. \textbf{Line-Path Search} asks about the spatial relation between candidate objects and the finite line segment based on a question at a queried sound event moment, and \textbf{Trajectory Distance} asks about the observed trajectories of agents, \textit{Ego} or \textit{Exo}, over a temporal interval between two sound events, respectively (Figure\,\ref{fig:stage_task_guide}).

\begin{itemize}[leftmargin=*, itemsep=0pt]
    \item \textbf{Line-Path Search (static).} The question asks which candidate object is closest to the straight-line segment connecting the camera wearer to a destination static object explicitly mentioned in the question at a sound event moment.

    \item \textbf{Line-Path Search (dynamic).} The question asks which candidate object is closest to the straight-line segment connecting the camera wearer to the other dynamic agent's position at a sound event moment.

    \item \textbf{Trajectory Distance.} The question asks how far both the camera wearer and the other agent actually travel between two sound events, where the trajectories are affected by physical obstacles rather than reflecting the displacement between the start and end positions.
\end{itemize}

\paragraph{Ground-truth geometry calculation.}
\textit{Line-Path Search} compares the candidate object positions listed in the question with the finite segment between camera wearer and target destination. For endpoints $\mathbf{a}$ and $\mathbf{b}$ and a candidate object position $\mathbf{o}$, its distance to the segment is
\begin{equation}
    d(\mathbf{o}, [\mathbf{a},\mathbf{b}])
    = \min_{\lambda \in [0,1]}
    \left\|\mathbf{o} - \bigl(\mathbf{a} + \lambda(\mathbf{b}-\mathbf{a})\bigr)\right\|_2.
    \label{eq:appendix_line_path_distance}
\end{equation}
In both \textit{Line-Path Search} tasks, $\mathbf{a}$ is the camera wearer's position ($\mathbf{p}_{w}$) at the sound event query time. For the \textbf{static}, $\mathbf{b}$ is the position of the destination object ($\mathbf{p}_{obj}$) described in the question. For the \textbf{dynamic}, $\mathbf{b}$ is the other agent's position ($\mathbf{p}_{o}$) at the same query time. The answer is the candidate object with the smallest distance to this segment, which need not lie exactly on the hypothetical segment.

For \textit{Trajectory Distance}, the ground truth is a pair of macro-scale path lengths for the camera wearer and the other agent within the interval between the two queried sound events. To reduce error accumulation from localization jitter and small body movements, we apply a 0.75\,s moving average to the trajectory within the interval, followed by Ramer--Douglas--Peucker (RDP) polyline simplification with a tolerance of 0.5\,m, chosen to reduce the effect of small local movements, such as taking a step forward and back, while retaining the macro-scale route. The ground-truth distance is the sum of segment lengths along this smoothed and simplified path. Let $\mathbf{q}_{a,0},\ldots,\mathbf{q}_{a,K_a}$ denote the $K_a+1$ retained vertices for agent $a\in\{\mathrm{ego},\mathrm{other}\}$. Then
\begin{equation}
    L_a = \sum_{k=1}^{K_a}\left\|\mathbf{q}_{a,k}-\mathbf{q}_{a,k-1}\right\|_2,
    \qquad \mathbf{y}_{\mathrm{GT}}=(L_{\mathrm{ego}},L_{\mathrm{other}}).
    \label{eq:appendix_trajectory_length}
\end{equation}

The model predicts both distances in meters, ordered as the camera wearer first and the other agent second. The task score is the average of the two agents' AbsMRA scores.

\paragraph{Dataset statistics.}
The \textit{path reasoning} category contains 596 QAs: 196 \textit{Line-Path Search (static)}, 200 \textit{Line-Path Search (dynamic)}, and 200 \textit{Trajectory Distance} questions. Both \textit{Line-Path Search} variants use multiple-choice answers evaluated by exact-match accuracy. \textit{Trajectory Distance} requires a numerical answer in meters and uses AbsMRA. The overall \textit{path reasoning} score is the average of these three task scores (Table\,\ref{tab:ourbench}).

\section{Comparison with the SAVVY Framework}

SAVVY~\cite{chen2025savvy} introduces a pioneering training-free framework for 3D spatial reasoning in dynamic audio-visual environments, but it has several conceptual and structural limitations by strictly decoupling spatial computations from the foundation model.
To overcome these inherent bottlenecks, \ourmodel explicitly injects global spatial geometry directly into the AV-LLM via a synchronized 2D floormap as a visual input.
In the following, we compare the two frameworks from multiple perspectives (Figure\,\ref{fig:framework_comparison}) to describe their design differences and highlight the benefits of \ourmodel{} for multimodal spatial reasoning.

\subsection{Utilization of AV-LLMs}

\begin{itemize}[leftmargin=*, itemsep=0pt]
    \item \textbf{SAVVY: Underutilizing multimodal reasoning capacity.}
    The SAVVY framework strictly utilizes the AV-LLM as an initial preprocessing tool rather than a core reasoner. The model is queried to generate ``snapshot descriptors'' that identify when the target event occurred, what relevant objects are present, and spatial estimates from snapshots. All subsequent actual spatial processing, including distance and direction calculation procedures, is delegated to external, rule-based offline modules. These offline modules derive the final answer by mathematically computing the audio DoA via the SRP-PHAT algorithm~\cite{dibiase2000high}, performing visual segmentation, and applying rigid coordinate transformations. Consequently, the final spatial answer is determined entirely by mathematical operations outside the model.

    \item \textbf{\ourmodel: Maximizing multimodal reasoning capacity.}
    In contrast, our framework proposes a fundamental reversal of the SAVVY paradigm. Instead of calculating spatial answers after the AV-LLM, we explicitly render critical spatial information, including the camera pose, acoustic DoA, and 3D point cloud boundaries, into a 2D floormap that serves as a direct visual prompt to the model. By feeding this integrated top-down map into the AV-LLM alongside the egocentric audio-visual streams, we let the model fully use its cross-modal intelligence.
\end{itemize}

\subsection{Information Processing Pipeline}

\begin{itemize}[leftmargin=*, itemsep=0pt]
    \item \textbf{SAVVY: Multi-stage pipeline.}
    SAVVY relies on a multi-stage architecture that extracts visual and acoustic spatial cues through separate modules.
    After generating an initial text-based scene summary, these textual descriptions then guide external vision models (\ie CLIPSeg~\cite{luddecke2022image}, SAM2~\cite{ravi2025sam}, and a monocular metric depth estimator~\cite{bhat2023zoedepth}) and discrete acoustic algorithms (\ie CDR~\cite{schwarz2015coherent}, SRP-PHAT~\cite{dibiase2000high}) to independently extract spatial cues.
    Finally, these disjointed signals are clustered~\cite{khan2014dbscan} and fused via a Kalman filter and offline mathematical operations to compute the ultimate spatial answer.

    \item \textbf{\ourmodel: Single inference.}
    We synchronize the pre-rendered 2D floormap video with the standard egocentric RGB footage and directly inject them into the model as dual-stream videos, alongside the audio track and a text prompt including the question and floormap guide.
    After the semantic grounding is ready for the scene, the AV-LLM performs complex cross-modal reasoning for any spatial-aware task with a single inference.
\end{itemize}

\subsection{Future Scalability}

\begin{itemize}[leftmargin=*, itemsep=0pt]
    \item \textbf{SAVVY: Marginal benefit from model advancements.}
    In SAVVY, the AV-LLM is relegated to performing simple, auxiliary sub-tasks, such as temporal grounding and bounding box detection.
    Because the core solving procedure, including geometric transformations, is entirely offloaded to external, rule-based mathematical modules, the SAVVY pipeline still relies on the AV-LLM's grounding capabilities from the egocentric view, even as the spatial intelligence of future AV-LLMs improves.

    \item \textbf{\ourmodel: Scaling with model evolution.}
    Our framework treats the AV-LLM as the central spatial reasoner, directly feeding it the 2D floormap to deduce geometric relationships. While the map reading and spatial reasoning capabilities of current models are not yet at an ideal level, our direct-injection approach ensures that the system's overall performance is directly coupled with the model's own capabilities. As the intrinsic spatial intelligence of foundation models continues to evolve, \ourmodel{} can benefit directly from these advances.

\end{itemize}

\begin{figure}[!t]
    \centering
    \includegraphics[width=\linewidth]{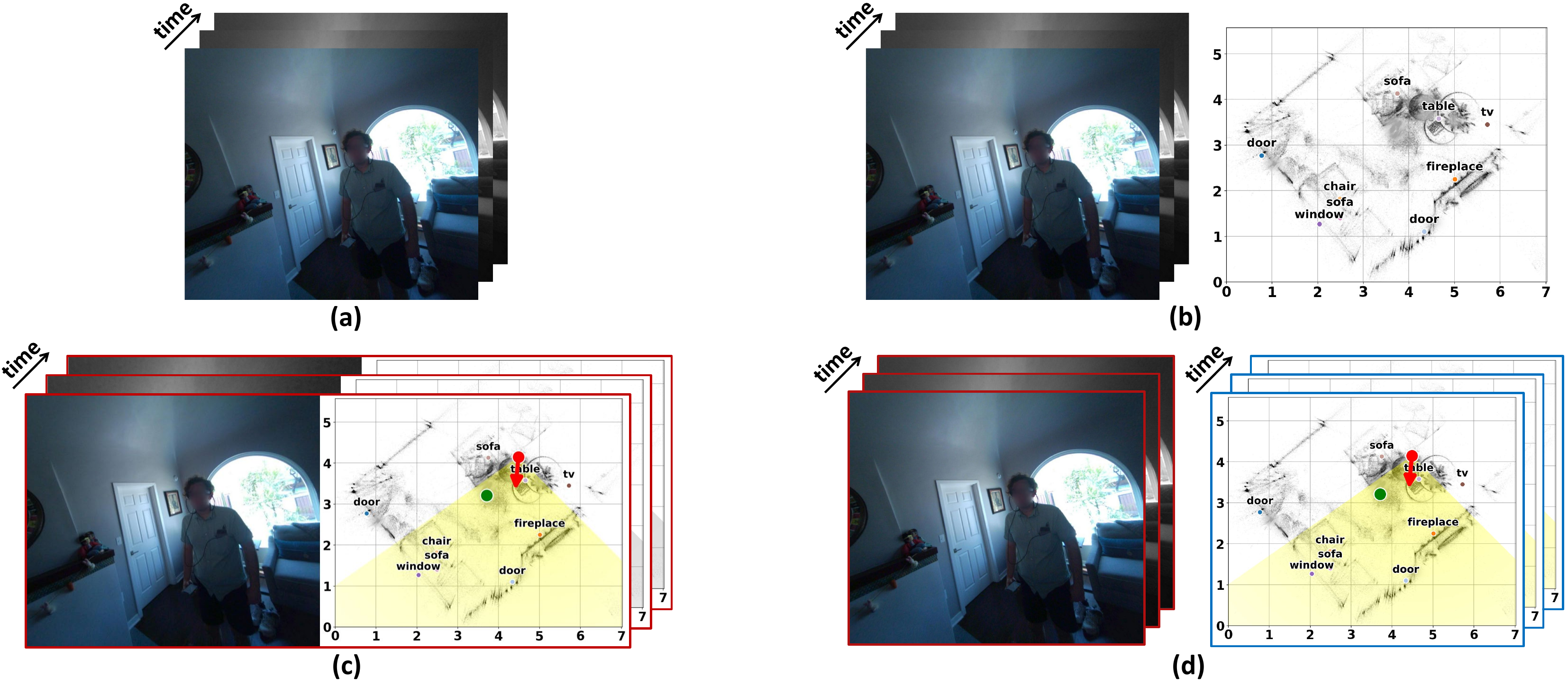}
    \vspace{-20pt}
    \caption{
    \textbf{Comparison of visual input formats for floormap prompting.} The strategies include (a) the RGB-only baseline, (b) adding a static floormap image, (c) merging both views into a single-stream side-by-side video, and (d) our proposed dual-stream injection using separate synchronized videos.
    }
    \label{fig:floormap_prompting}
    \medskip
        \small
    \captionsetup{type=table}
    \caption{\textbf{Comparison of floormap prompting methods for \ourmodel.} All experiments are evaluated using the Qwen3-Omni-30B model. Dual-stream denotes providing the egocentric RGB footage and the rendered 2D floormap as two separate, synchronized video inputs to the AV-LLM. For each input configuration, refer to Figure\,\ref{fig:floormap_prompting}.
    }
    \label{tab:floormap_prompting}
    \vspace{5pt}
    \addtolength{\tabcolsep}{-1pt}
    \resizebox{\textwidth}{!}{%
    \begin{tabular}{ll|ccc|ccc|c}
        \toprule
        & & \multicolumn{3}{c|}{Egocentric} & \multicolumn{3}{c|}{Exocentric} & \\
        \cmidrule(lr){3-5} \cmidrule(lr){6-8}
        Method & Configuration & \!\!Dir\,(Simple)\!\! & \!\!Dir\,(Hard)\!\! & Dist & \!\!Dir\,(Simple)\!\! & \!\!Dir\,(Hard)\!\! & Dist & Overall \\
        \midrule
        RGB-only & Figure\,\ref{fig:floormap_prompting}(a)                  & 72.7 & 70.7 & 59.4 & 34.1 & 24.1 & 31.8 & 48.1 \\
        + Floormap (Image) & Figure\,\ref{fig:floormap_prompting}(b)        & 76.1 & 71.6 & 57.7 & 28.8 & 26.6 & 30.1 & 47.3 \\
        + Floormap (Single-stream) & Figure\,\ref{fig:floormap_prompting}(c)& 68.1 & 58.2 & 57.5 & 27.8 & 31.6 & 25.6 & 44.0 \\
        + Floormap (Dual-stream) & Figure\,\ref{fig:floormap_prompting}(d) & 74.4 & 71.1 & 61.8 & 31.6 & 34.4 & 31.0 & 49.6 \\
        \bottomrule
    \end{tabular}%
    }

\end{figure}

\subsection{Task Flexibility and Broad Applicability}

\begin{itemize}[leftmargin=*, itemsep=0pt]
    \item \textbf{SAVVY: Task-specific rigidity.}
    Due to its strict algorithmic design, SAVVY requires a highly specific question structure to function properly.
    It strictly requires the precise occurrence of a distinct sound event and the explicit presence of target objects to execute its snapshot description. This rigidity limits the flexibility of the framework for broader spatial questions that lack these exact geometric anchors.

    \item \textbf{\ourmodel: Task flexibility.}
    By contrast, our framework leverages the generalized visual reasoning capabilities of MLLMs, making it structurally flexible.
    Because \ourmodel fundamentally provides a rich spatial hint via the top-down floormap, it is not restricted to rigid, template-based question formats.
    Beyond SAVVY-Bench, \ourbench spans various tasks associated with cross-agent, semantic region, and physical object concepts, which \ourmodel can address but SAVVY is inherently restricted from.
\end{itemize}

\section{Additional Results}
\label{app:additional_results}

\subsection{Floormap Prompting Strategies}
\label{app:prompting_strategies}

To determine the most effective way to inject the rendered 2D floormap into AV-LLMs, we evaluate four different prompting strategies, as shown in Figure\,\ref{fig:floormap_prompting}.
The four strategies are: (1) RGB-only, the baseline using only the egocentric video; (2) + Floormap (Image), which adds a single static 2D map without temporal trajectories or dynamic sound information; (3) + Floormap (Single-stream), which concatenates the egocentric and floormap videos side-by-side into a single dense video; and (4) + Floormap (Dual-stream), which feeds the two videos as separate, synchronized inputs.

Table\,\ref{tab:floormap_prompting} reports the per-task breakdown of these prompting strategies. Our final \ourmodel framework adopts the dual-stream strategy, which reaches the highest overall performance (49.6\%) by letting the model cross-reference the egocentric appearance against the dynamic map geometry without visual interference.

\paragraph{Floormap and prompting ablations.}
Floormap injection methodology matters more than whether it is present. With Qwen3-Omni-30B, a static floormap image gives no benefit over RGB-only (47.3\% vs.\ 48.1\%), since it carries neither the wearer's motion nor the moving sound source. Side-by-side concatenation into one video is worse (44.0\%): the egocentric half always occupies the left of the frame, and the model becomes biased toward answering ``\textit{left}''. The dual-stream form reaches 49.6\% on identical information, a +5.6\% gap from layout alone. Semantic grounding adds a further +4.3\% on Gemini-2.5-Flash and +5.1\% on Pro (Appendix\,\ref{app:semantic_grounding}), concentrated in the \textit{exocentric} direction task.

Moreover, current MLLMs and AV-LLMs often struggle with processing excessively long or visually dense video inputs~\cite{wu2021towards, mangalam2023egoschema, weng2024longvlm, geng2025longvale, wang2025lvbench, shu2025video}, the same context length issue we further discuss in Appendix~\ref{app:failure_analysis_2}. Because of this limitation, feeding both videos overwhelms the model and can degrade the overall performance in simple tasks. On the other hand, the static floormap image avoids this heavy video burden and performs surprisingly well on a relatively simpler task, \textit{egocentric} direction (76.1\% on Simple, 71.6\% on Hard). However, because the static image lacks temporal changes and dynamic sound source localization, it fails to generalize to harder tasks like distance estimation and exocentric reasoning.

\begin{table}[!t]
    \centering
    \small
    \caption{\textbf{Ablation study on input video frame counts.} The frame count indicates the number of uniformly sampled frames from each input video stream, an egocentric view and a floormap.}
    \label{tab:frames_number}
    \vspace{-5pt}
    \begin{tabular}{c|cc|cc|c}
        \toprule
        & \multicolumn{2}{c|}{Egocentric} & \multicolumn{2}{c|}{Exocentric} & \\
        \cmidrule(lr){2-3} \cmidrule(lr){4-5}
        \# of Frames & Direction & Distance & Direction & Distance & Overall \\
        \midrule
        32  & 73.9 & 63.9 & 31.6 & 31.0 & 50.1 \\
        64  & 70.6 & 62.8 & 31.9 & 28.9 & 48.6 \\
        128 & 72.8 & 61.8 & 32.9 & 31.0 & 49.6 \\
        \bottomrule
    \end{tabular}
\end{table}

\subsection{Number of Video Frames}
Table\,\ref{tab:frames_number} presents results depending on the number of sampled video frames used in our dual-stream injection. We observe that utilizing 32 frames achieves the highest overall accuracy (50.1\%), particularly excelling in \textit{egocentric} tasks with 73.9\% in direction and 63.9\% in distance. This finding directly aligns with the context length bottleneck of current AV-LLMs discussed earlier. Nonetheless, we adopt 128 frames as our default configuration to prioritize complex exocentric tasks, which require tracking dynamic sound sources and relating them to reference objects over time.

\begin{table}[!t]
    \centering
    \small
    \caption{\textbf{Ablation study on semantic grounding with Gemini-2.5-Flash and Pro on SAVVY-Bench.} The (--) Semantic Grounding setting evaluates our \ourmodel framework without explicit object labels overlaid on the rendered 2D floormap on SAVVY-Bench.}
    \label{tab:gemini_flash}
    \vspace{-5pt}
    \addtolength{\tabcolsep}{1pt}
    \resizebox{\textwidth}{!}{
    \begin{tabular}{l|cc|cc|c}
        \toprule
        & \multicolumn{2}{c|}{Egocentric} & \multicolumn{2}{c|}{Exocentric} & \\
        \cmidrule(lr){2-3} \cmidrule(lr){4-5}
        Method & Direction & Distance & Direction & Distance & Overall \\
        \midrule
        Gemini-2.5-Flash~\cite{comanici2025gemini} & \textbf{74.2} & 49.7 & 29.8 & 29.0 & 45.7 \\
        Gemini-2.5-Flash-\textbf{\ourmodel} & {71.3} & 55.4 & \textbf{44.5} & \textbf{32.2} & \textbf{50.9} \\
        {(--) Semantic Grounding} & 63.3 & \textbf{58.0} & 34.4 & 30.8 & 46.6 \\
        \midrule
        Gemini-2.5-Pro~\cite{comanici2025gemini} & 75.2 & \textbf{59.6} & 31.7 & \textbf{37.0} & 50.9 \\
        Gemini-2.5-Pro-\textbf{\ourmodel} & \textbf{75.8} & 55.4 & \textbf{52.8} & 34.9 & \textbf{54.7}  \\
        {(--) Semantic Grounding} & 77.6 & 56.0 & 35.6 & 29.4 & 49.6 \\
        \bottomrule
    \end{tabular}
    }
\end{table}

\begin{table}[t]
    \centering
    \small
    \captionsetup{font=footnotesize}
    \caption{\textbf{Ablation study on semantic grounding with Gemini-3.6-Flash on \ourbench.}
    The (–) Semantic Grounding setting evaluates \ourmodel framework without explicit object labels overlaid on the rendered 2D floormap on \ourbench.
    $^*$Pooled Visit Order and Unvisited questions.}
    \label{tab:sem_gnd_ablation_saved}
    \vspace{-5pt}
    \renewcommand{\arraystretch}{0.95}
    \setlength{\tabcolsep}{5pt}
    \begin{tabular}{l|ccc}
        \toprule
        Task
        & \shortstack{Gemini-3.6-Flash}
        & \shortstack{Gemini-3.6-Flash-\\\textbf{FloorSAV}}
        & ($-$) Semantic Grounding \\
        \midrule
        \rowcolor{gray!15}
        \multicolumn{4}{l}{\textcolor{RoyalBlue}{\textbf{Dynamic Relativity}}} \\
        \hspace{1em}Ego-to-Exo Direction
        & 56.0 & \textbf{65.0} & 59.5\loss{5.5} \\
        \hspace{1em}Ego-to-Exo Distance (AbsMRA)
        & 34.3 & \textbf{34.6} & 32.1\loss{2.5} \\
        \hspace{1em}Exo-to-Ego Direction
        & 32.0 & \textbf{50.5} & 47.5\loss{3.0} \\
        \hspace{1em}Exo-to-Ego Distance (AbsMRA)
        & 46.4 & \textbf{44.7} & 41.1\loss{3.6} \\
        \hspace{1em}\textit{4-task average}
        & 42.2 & \textbf{48.7} & 45.0\loss{3.7} \\
        \midrule
        \rowcolor{gray!15}
        \multicolumn{4}{l}{\textcolor{DarkOrange}{\textbf{Regional}}} \\
        \hspace{1em}Location Awareness
        & 92.0 & \textbf{94.5} & 92.5\loss{2.0} \\
        \hspace{1em}Visit History$^*$
        & 86.7 & \textbf{92.3} & 91.8\loss{0.5} \\
        \hspace{1em}\textit{2-task average}
        & 89.4 & \textbf{93.4} & 92.2\loss{1.2} \\
        \midrule
        \rowcolor{gray!15}
        \multicolumn{4}{l}{\textcolor{ForestGreen}{\textbf{Path Reasoning}}} \\
        \hspace{1em}Line-Path Search (static)
        & 60.2 & \textbf{77.0} & 55.1\loss{21.9} \\
        \hspace{1em}Line-Path Search (dynamic)
        & 73.0 & \textbf{78.0} & 77.5\loss{0.5} \\
        \hspace{1em}Trajectory Distance (AbsMRA)
        & 16.9 & 25.0 & \textbf{25.4}\gain{0.4} \\
        \hspace{1em}\textit{3-task average}
        & 50.0 & \textbf{60.0} & 52.7\loss{7.3} \\
        \midrule
        \textbf{Overall (QA-weighted)}
        & 58.3 & \textbf{65.3} & 61.3\loss{4.0} \\
        \bottomrule
    \end{tabular}%
\end{table}

\subsection{Semantic Grounding Effect}
\label{app:semantic_grounding}
Tables \,\ref{tab:gemini_flash} and \,\ref{tab:sem_gnd_ablation_saved} show an ablation study on the role of semantic grounding. In our default \ourmodel framework, the respective base models autonomously generate explicit object labels overlaid on the 2D map. Because this process relies on the model's own visual reasoning capabilities, we observe distinct behaviors between the Gemini-2.5-Pro and Flash models on SAVVY-Bench, and Gemini-3.6-flash on SAVED-Bench.

Omitting the semantic grounding process degrades overall performance for all cases, which highlights the role of accurate object labels for spatial reasoning. Especially for tasks that require explicit environmental context, \textit{Exocentric} direction (Table \,\ref{tab:gemini_flash}) and \textit{Line-Path Search (static)} (Table\,\ref{tab:sem_gnd_ablation_saved}), the decline in performance is substantial without semantic grounding.

\section{Qualitative Examples}
\label{app:qualitative}

In this section, we qualitatively evaluate the spatial reasoning capabilities of our \ourmodel framework using the Gemini series. We first present successful cases where the injected 2D floormap effectively guides the model to solve complex tasks. Following this, we categorize and analyze common failure modes into three distinct parts to provide deeper insights into the remaining cognitive bottlenecks of current foundation models.

\subsection{Successful Examples on SAVVY-Bench}
\label{app:success_examples_savvy}

We demonstrate the effectiveness of \ourmodel through two challenging exocentric scenarios: direction and distance estimation. In the direction case, the model must deduce the sound source's location relative to a hypothetical robot's viewpoint. We utilize Gemini-2.5-Flash as the AV-LLM, following the evaluation protocol for SAVVY-Bench (Section~\ref{sec:experiments}).

\paragraph{How the AV-LLM reads the floormap.}
\label{subsec:qualitative}

\begin{figure}[!t]
    \centering
    \includegraphics[width=\linewidth]{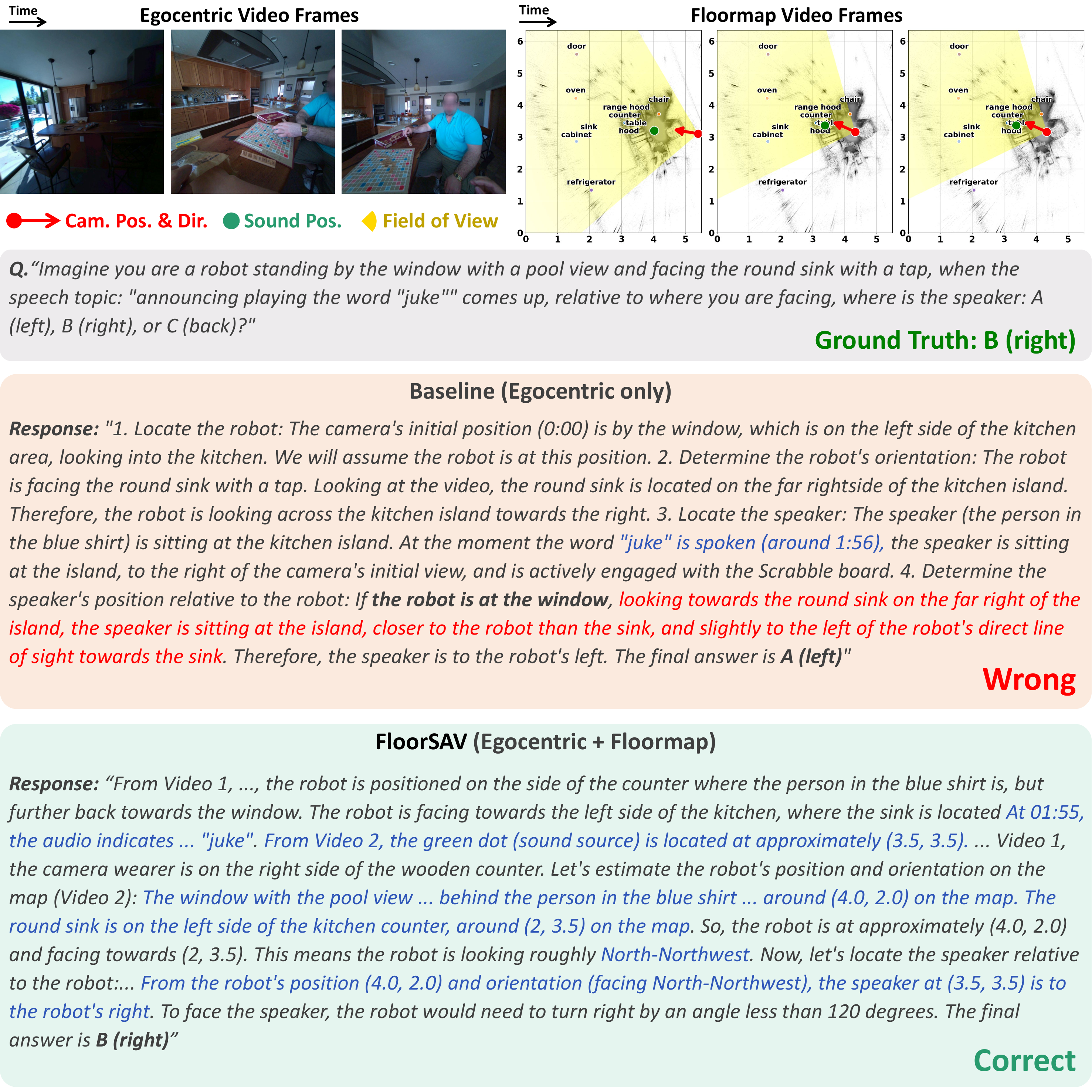}
    \vspace{-15pt}
    \caption{\textbf{Qualitative example of an exocentric direction question.} The baseline sees only the egocentric video and has to imagine the robot's viewpoint, answering ``\textit{left}''. \ourmodel reads the robot, the facing object, and the green sound-source marker off the floormap grid and answers ``\textit{right}''.}
    \label{fig:qualitative_allo_dir}
\end{figure}

Figure\,\ref{fig:qualitative_allo_dir} traces one \textit{exocentric} direction question through both models. A robot stands at the window facing the round sink, and the question asks where the speaker is at the moment a given word is spoken. The baseline has to imagine a viewpoint it never sees. It puts the sink on the far right of the kitchen island, concludes that the speaker falls left of that line of sight, and answers ``\textit{left}''. \ourmodel reads coordinates instead. It places the robot near $(4.0, 2.0)$ and the sink near $(2.0, 3.5)$, derives a north-northwest heading, and finds the green sound-source marker at $(3.5, 3.5)$ to the robot's right. The \textit{exocentric} gains in Table\,\ref{tab:savvy_benchmark} follow from this shift from visual imagination to arithmetic on an explicit frame.

\begin{figure}[!t]
    \centering
    \includegraphics[width=\linewidth]{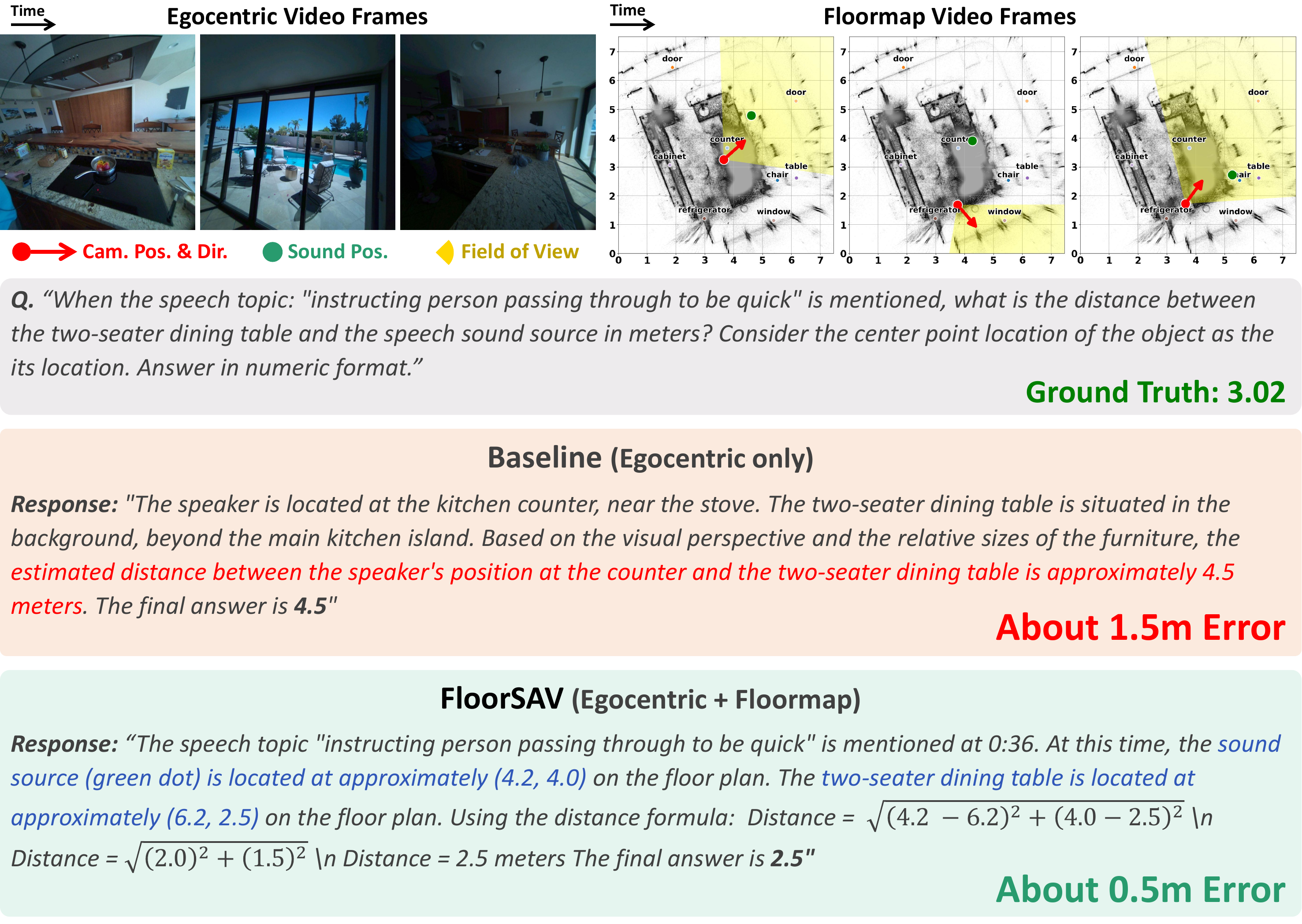}
    \vspace{-20pt}
    \caption{\textbf{Qualitative example of the \textit{exocentric distance} task.} The baseline model suffers from severe depth distortion inherent in monocular egocentric vision, overestimating the metric distance by approximately 1.5 meters. In contrast, \ourmodel extracts precise metric coordinates from the top-down floormap to compute the Euclidean distance, significantly reducing the measurement error to about 0.5 meters.}
    \label{fig:appendix_allo_dist}
\end{figure}

Similarly, in the exocentric distance task, estimating exact metric distances between two physical locations is exceptionally difficult when relying only on egocentric visual perspectives. Consequently, in Figure\,\ref{fig:appendix_allo_dist}, the baseline model significantly overestimates the distance, resulting in an error of approximately 1.5 meters. However, by utilizing the synchronized floormap, \ourmodel accurately extracts the specific grid coordinates for both the \textit{sound source (4.2, 4)} and the target \textit{dining table (6.2, 2.5)} at the queried timestamp. By computing the Euclidean distance directly with these map coordinates, the model reduces the geometric error, arriving at a more accurate measurement of 2.5 meters. These examples clearly illustrate how explicitly injecting the global map geometry empowers the model to perform precise mathematical and spatial reasoning.

\begin{figure}[!t]
    \centering
    \includegraphics[width=\linewidth]{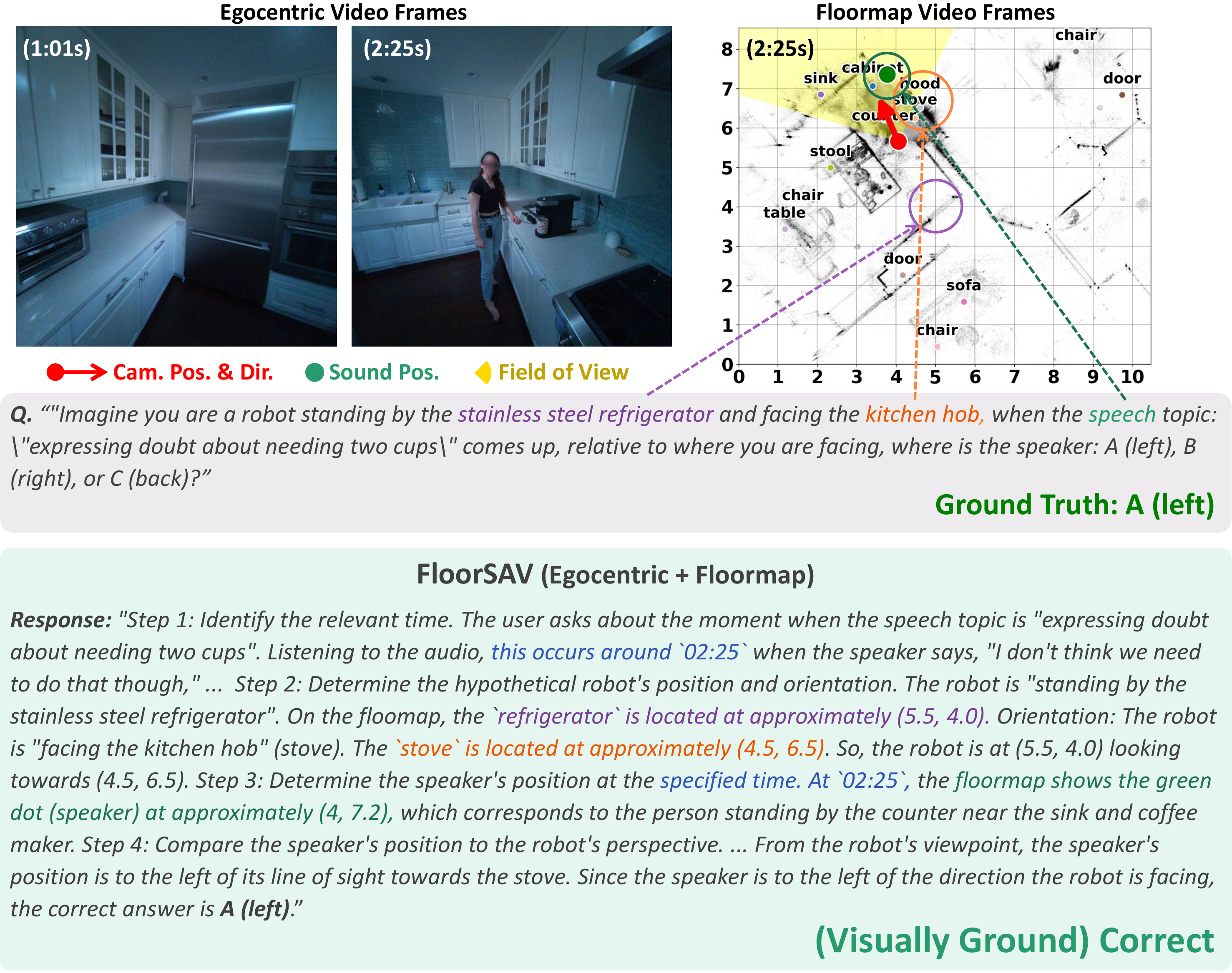}
    \vspace{-16pt}
    \caption{\textbf{Qualitative example of the semantically grounded floormap.} Our semantically grounded floormap, constructed through the proposed pipeline with Gemini, offers semantic labels as visually grounded cues that enhance the model's spatial reasoning.}
    \label{fig:appendix_visual_ground}
    \vspace{-10pt}
\end{figure}

Figure~\ref{fig:appendix_visual_ground} shows how the semantic labels help. The AV-LLM reads the coordinates of the task-related objects from the labels, the \textit{refrigerator (5.5, 4)} (slightly off) and \textit{kitchen hob (4.5, 6.5)}, locates the speaker at \textit{(4, 7.2)}, and answers the \textit{exocentric} direction question correctly. Errors from the object visual mapping stage carry over to the floormap, but the model still reasons through them to the right answer.

\subsection{Successful Examples on \ourbench}
\label{app:success_examples_savved}

We further demonstrate the benefits of \ourmodel{} on SAVED-Bench through one qualitative example for each category: \textit{dynamic relativity}, \textit{regional}, and \textit{path reasoning} QAs, which highlight cross-agent viewpoint transformation, semantic region recognition, and path-aware reasoning about physical surroundings, respectively. We utilize Gemini-3.6-Flash as the AV-LLM, following the evaluation protocol for \ourbench (Section~\ref{sec:experiments}).

\begin{figure}[!t]
    \centering
    \includegraphics[width=\linewidth]{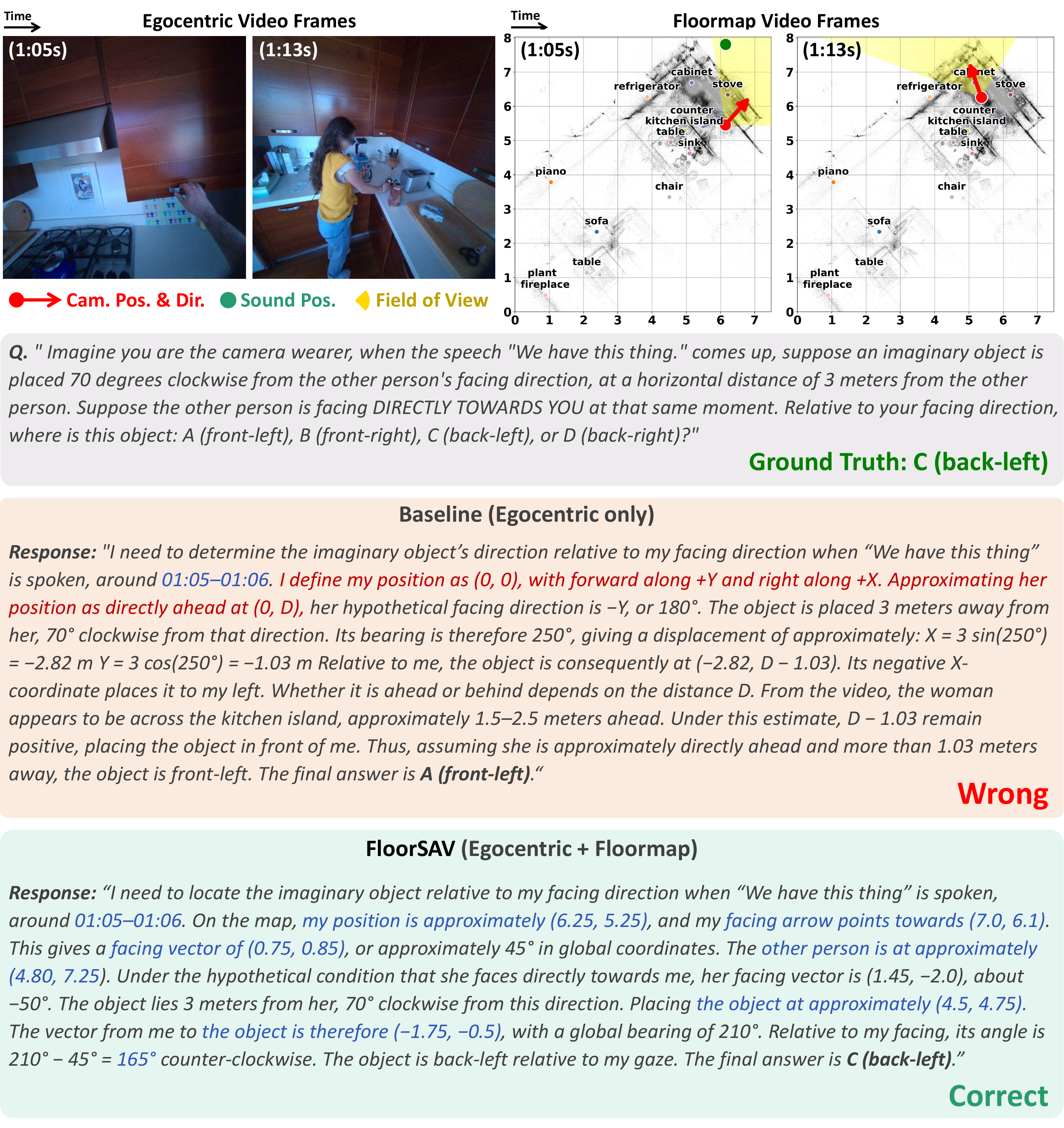}
    \vspace{-20pt}
    \caption{\textbf{Qualitative example of dynamic relativity QA.}
    In this FoV-out \textit{Exo-to-Ego} direction example, both methods correctly identify the sound event time. However, the baseline incorrectly assumes that the other agent is directly ahead ($+y$) of the camera wearer. \ourmodel estimates the other agent's position from adjacent frames and utilizes the floormap as a shared global coordinate frame for viewpoint transformation.}
    \label{fig:appendix_saved_dynrel}
\end{figure}

Figure\,\ref{fig:appendix_saved_dynrel} shows an \textit{Exo-to-Ego} direction example in which the other agent is invisible to the camera at the querying moment (FoV-out). The baseline primarily constructs a hypothetical reference frame centered on the camera wearer; however, the model incorrectly grounds the other agent's position in its frame, which leads to a wrong answer. In contrast, \ourmodel{} uses floormap coordinates as a shared reference frame. Within this frame, the model estimates the other agent's position, computes the imaginary object's location, and finally determines its relative direction from the camera wearer's viewpoint. This example illustrates how explicit representation in a unified global frame supports the cross-agent viewpoint transformation.

\begin{figure}[!t]
    \centering
    \includegraphics[width=\linewidth]{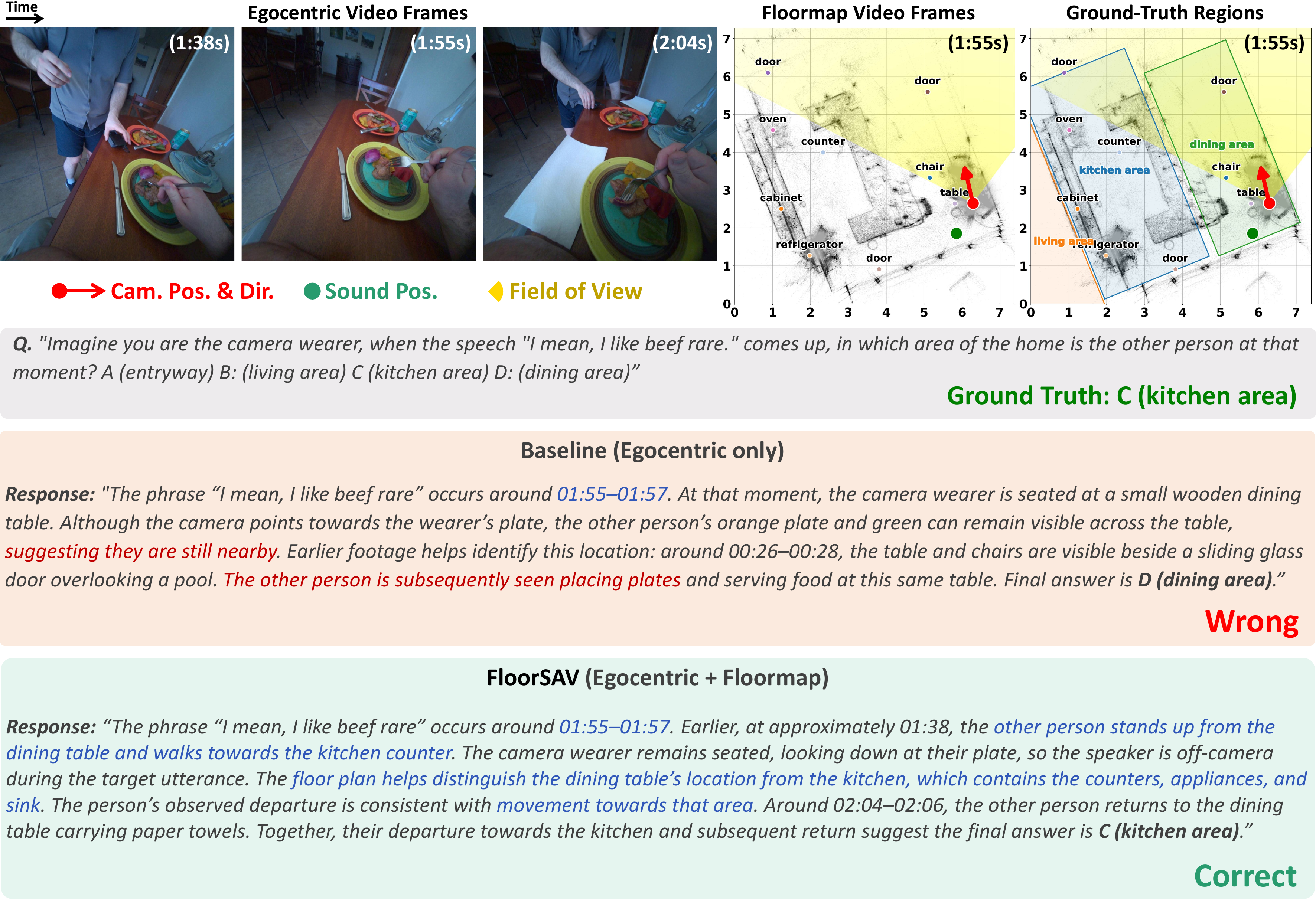}
    \vspace{-20pt}
    \caption{\textbf{Qualitative example of regional QA.}
    For the \textit{Location Awareness} task, both methods correctly identify the sound event time. However, the baseline incorrectly assumes that the other agent remains near the dining table. \ourmodel jointly reasons about the other agent's movement from nearby frame cues and physical layout on the floormap, identifies the region (kitchen area) from key objects in the floormap to answer correctly.}
    \label{fig:appendix_saved_regional}
\end{figure}

For the \textit{regional} QAs, specifically in \textit{Location Awareness} tasks (Figure\,\ref{fig:appendix_saved_regional}), the baseline heavily relies on an earlier frame showing the other agent near the dining table and assumes the other agent stays near the table. However, \ourmodel utilizes adjacent frames to infer the other agent's movement in relation to the physical surroundings represented on the floormap. It then identifies the region (kitchen area) based on key objects close to the other agent's estimated position. Through joint reasoning on the other agent's dynamics and the spatial arrangement of objects, \ourmodel successfully links the agent's estimated position to the correct semantic region.

\begin{figure}[!t]
    \centering
    \includegraphics[width=\linewidth]{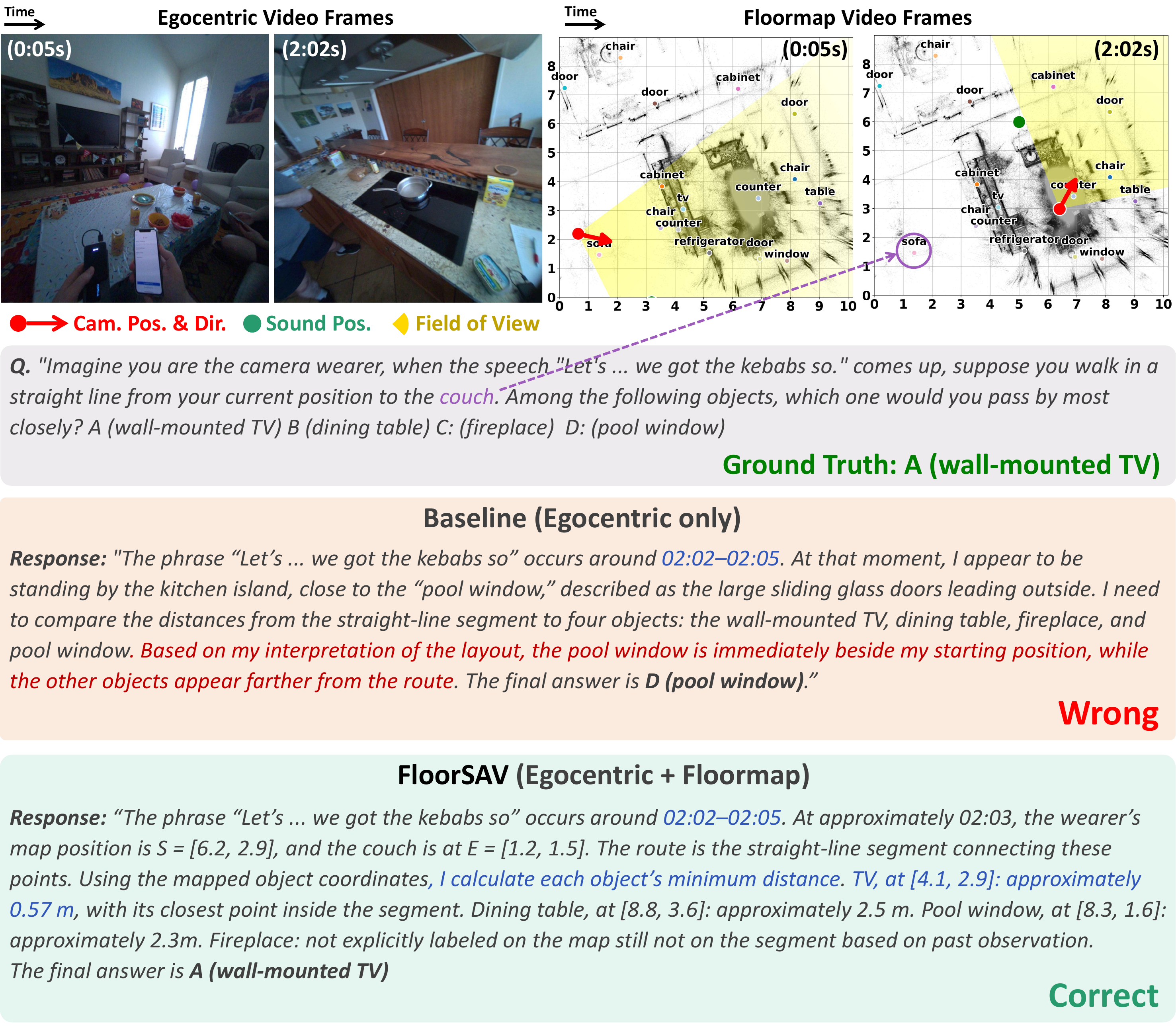}
    \vspace{-20pt}
    \caption{\textbf{Qualitative example of path reasoning QA.}
    In this \textit{Line-Path Search (static)} example, both methods correctly identify the sound event time. The baseline focuses on the nearby window and fails to relate a finite line segment to other candidates, which results in an inappropriate approach and answer. \ourmodel{} utilizes a 2D floormap as a reference frame and explicitly grounds the finite line segment and objects. With subsequent arithmetic calculations, \ourmodel determines the closest object to the finite line segment.}
    \label{fig:appendix_saved_path}
\end{figure}

In the \textit{Line-Path Search (static)} example from the \textit{path reasoning} QAs (Figure\,\ref{fig:appendix_saved_path}), the baseline focuses on the window visible in nearby frames and judges distances of remaining objects by their proximity to an inferred walking route, not from a finite line segment. This fails to present spatial reasoning on finite line segment and objects. By comparison, \ourmodel{} utilizes floormap coordinates to ground both the segment from the camera wearer's position to the target object and the candidate objects. Then the model computes the minimum distance from each candidate object to the segment in this shared global frame. This example effectively demonstrates how the 2D floormap provides shared spatial reference for physical objects and hypothetical paths, including objects outside the camera's current FoV.

\subsection{Failure Analysis 1: Temporal Localization Error}
\label{app:failure_analysis_1}

\begin{figure}[!t]
    \centering
    \includegraphics[width=\linewidth]{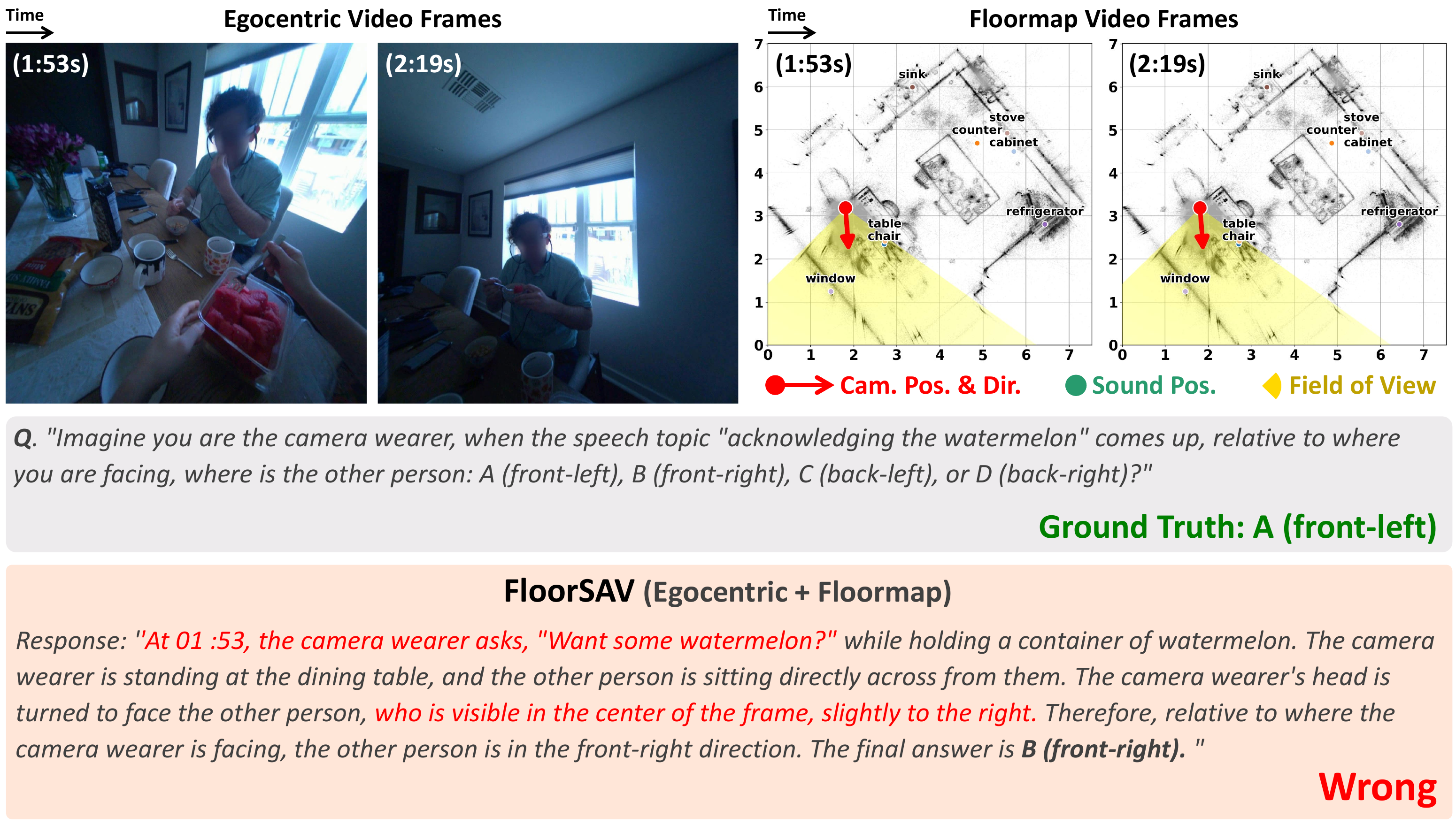}
    \vspace{-20pt}
    \caption{\textbf{Failure case analysis: Temporal localization error.} The model incorrectly points out the queried speech event (``\textit{acknowledging the watermelon}'', occurred at 02:19) to an earlier, incorrect timestamp (01:53). Consequently, it retrieves the spatial geometry from the wrong video frame, leading to a flawed directional prediction despite having access to the correct mapping mechanism.}
    \label{fig:appendix_temporal_error}
\end{figure}

The first common failure mode stems from the foundation model's inability to accurately pinpoint the precise time span of a queried audio-visual event. In highly dynamic environments, objects and camera wearers are constantly in motion. If the model anchors its reasoning to an incorrect timestamp, it inevitably retrieves the scene's spatial geometry from an irrelevant video frame. Consequently, even if \ourmodel successfully reads the map and processes the geometric calculations for that specific moment, the final prediction will be fundamentally flawed due to the incorrect temporal premise.

For instance, as illustrated in Figure\,\ref{fig:appendix_temporal_error}, the query asks for the relative direction of the speaker when the speech topic ``\textit{acknowledging the watermelon}'' occurs. Instead of grounding this event to the correct ground-truth timestamp (02:19), the model mistakenly anchors it to an earlier, semantically similar conversation at 01:53 (``\textit{Want some watermelon?}''). Because the relative positions of the camera wearer and the speaker at 01:53 differ entirely from their positions at 02:19, the model confidently but incorrectly predicts ``\textit{front-right}'' rather than the true answer, ``\textit{front-left}''. This highlights that robust spatial reasoning in \ourmodel remains strictly bottlenecked by the base model's temporal localization capabilities.

\subsection{Failure Analysis 2: Long-Context Understanding}
\label{app:failure_analysis_2}

\begin{figure}[!t]
    \centering
    \includegraphics[width=\linewidth]{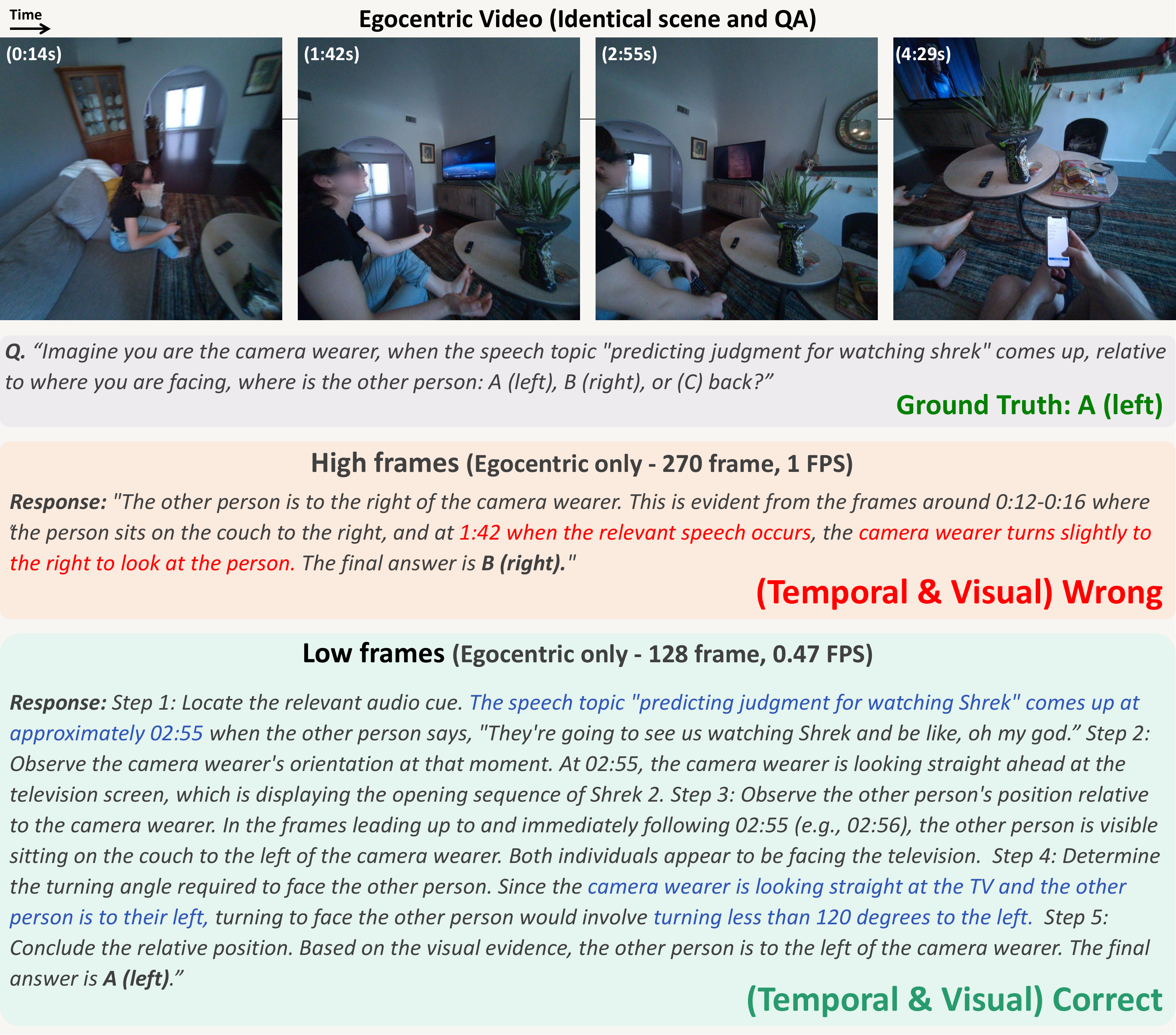}
    \vspace{-20pt}
    \caption{\textbf{Failure case analysis: Context length issue in the base model.} We compare high-density (270 frames, 1 FPS) and low-density (128 frames, 0.5 FPS) egocentric inputs. Contrary to the expectation that more frames provide better visual cues, the excessively long 270-frame input overwhelms the model and leads to an incorrect prediction, whereas the shorter 128-frame context successfully guides the model to the correct answer.}
    \label{fig:appendix_context_length_flash_vs_flash}
\end{figure}
\begin{figure}[!t]
    \centering
    \includegraphics[width=\linewidth]{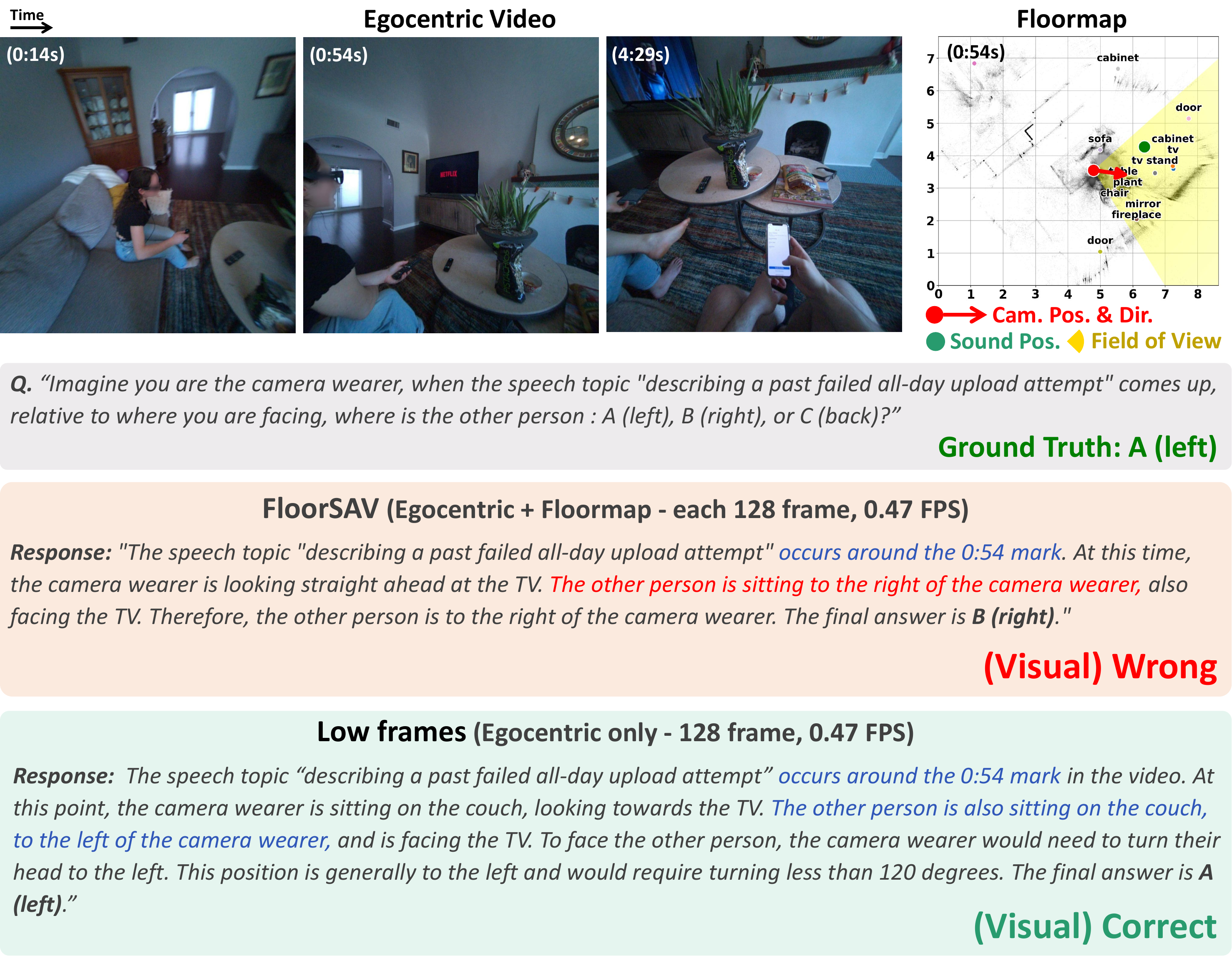}
    \vspace{-20pt}
    \caption{\textbf{Failure case analysis: Context length issue in the \ourmodel framework.} As \ourmodel doubles the input video tokens by incorporating the floormap stream, the model fails to process the basic RGB view correctly, inexplicably hallucinating the other person's location as ``\textit{right}'' despite them sitting on the left for almost the entire timeline.}
    \label{fig:appendix_context_length_flash_vs_floorSAV}
\end{figure}

The second major failure mode stems from the inherent context length limitations of current foundation models when processing long-form videos. Intuitively, one might assume that providing a higher frame rate would yield richer visual details and consequently improve spatial reasoning. However, our qualitative analysis reveals the opposite. As demonstrated in Figure\,\ref{fig:appendix_context_length_flash_vs_flash}, feeding the baseline model a dense 270-frame video (at 1 FPS) actually degrades its predictive capabilities, causing it to fail on a relatively simple \textit{egocentric} direction task. Conversely, reducing the input to a sparser 128 frames (approximately 0.5 FPS) significantly reduces the cognitive load, allowing the model to correctly identify the target. This confirms that excessively long video token sequences overwhelm the model's visual processing capacity, directly aligning with the findings from our frame count ablation study (Table\,\ref{tab:frames_number}).

This context length bottleneck becomes particularly critical when deploying our \ourmodel framework. Because \ourmodel injects the 2D floormap as an additional, synchronized video stream, it approximately doubles the total number of visual tokens fed into the base model. While this dual-stream approach provides essential geometric hints, it can overload the model that lacks robust long-context understanding. As shown in Figure\,\ref{fig:appendix_context_length_flash_vs_floorSAV}, the failure in \ourmodel does not arise from the model misunderstanding the floormap geometry. Rather, it shows a fundamental breakdown in basic visual perception. The model misreads the RGB view itself, incorrectly predicting that the person is on the ``\textit{right}'', although the individual is clearly seated on the left side of the screen for the vast majority of the timeline. This highlights that the successful application of \ourmodel heavily depends on the base model's capacity to handle massive multi-modal context without suffering from visual hallucination.

\subsection{Failure Analysis 3: Metric Map Misinterpretation}
\label{app:failure_analysis_3}

\begin{figure}[!t]
    \centering
    \includegraphics[width=\linewidth]{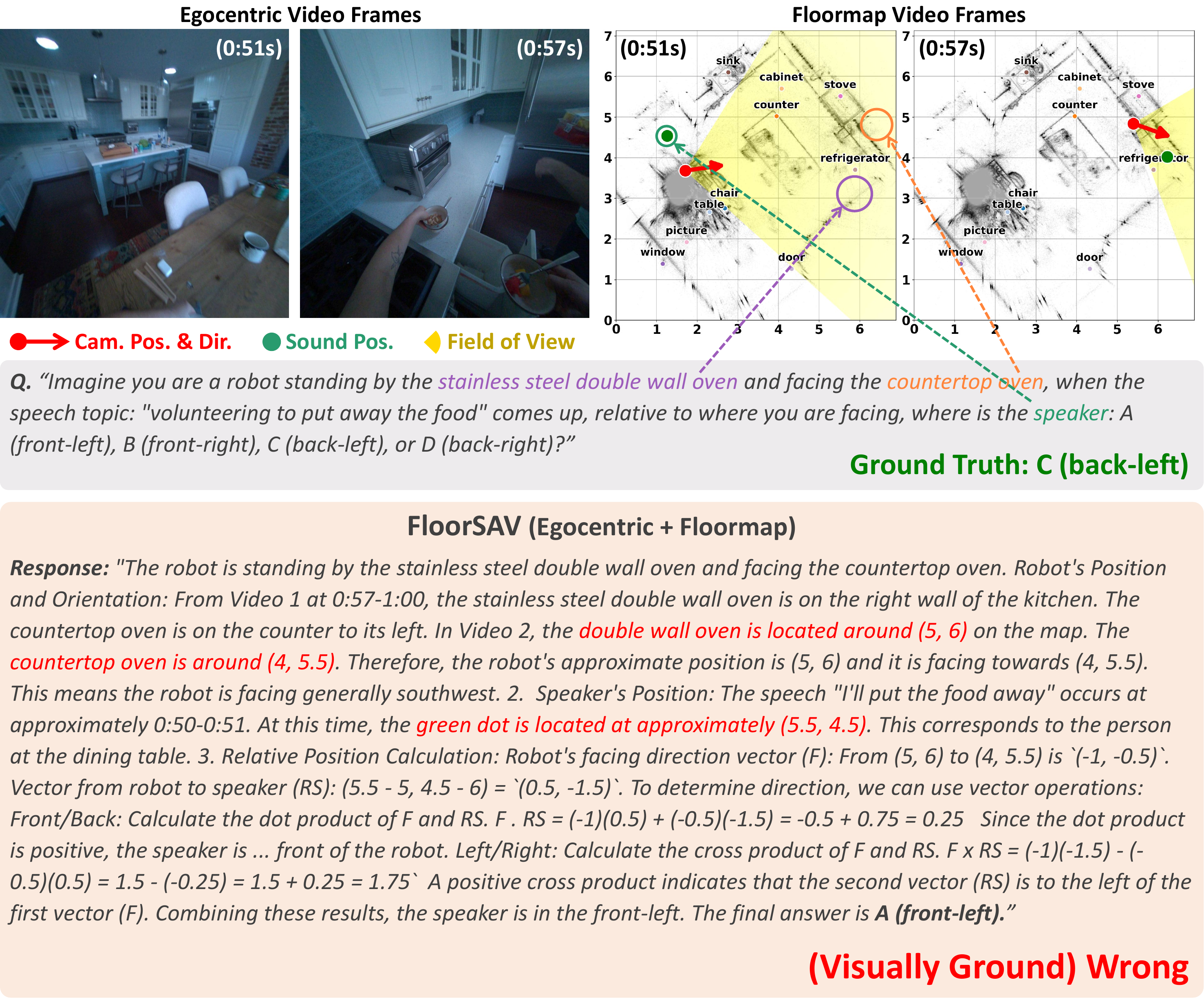}
    \vspace{-20pt}
    \caption{\textbf{Failure case analysis: Metric map misinterpretation.} Although \ourmodel accurately localizes the temporal window of the queried event, it misreads the 2D coordinate grid. It incorrectly specifies the locations of the sound source and reference objects, leading to wrong spatial calculation.}
    \label{fig:appendix_floormap_misunderstanding}
\end{figure}

The final major failure mode highlights a fundamental cognitive limitation of current foundation models, that they cannot reliably interpret abstract 2D metric grids. Even when the model successfully localizes the correct temporal window of an event, it often fails to extract precise Cartesian coordinates from the injected floormap. As illustrated in Figure\,\ref{fig:appendix_floormap_misunderstanding}, while \ourmodel accurately grounds the event time \textit{(0:50-0:51)}, it completely hallucinates the map coordinates during its reasoning process. It mistakenly reads the green sound source dot at \textit{(5.5, 4.5)} instead of its actual location near (1, 4.5), and similarly misplaces the double wall oven while only roughly guessing the countertop oven's location.
Artificially increasing the size of the visual markers on the map does not resolve this issue. This indicates that current MLLMs lack human-level proficiency in reading geometric grids and maps, likely due to a severe scarcity of top-down metric representations and coordinate-based spatial reasoning tasks in their training corpora. Consequently, explicit map reading remains a primary bottleneck for the \ourmodel framework.

\clearpage
\section{Floormap Guide Prompts}
\label{app:prompts}

Figure\,\ref{fig:sample_prompt} shows how the instruction prompt of the open-source model Qwen3-Omni-30B is assembled into the model's multimodal token sequence. Similarly, we construct two guide prompts for each video modality (egocentric and 2D floormap), which are provided to the Gemini series model. \texttt{Exp1} explains Video~1, while \texttt{Exp2} explains Video~2 in the dual-stream \ourmodel setting. The question and options are inserted immediately before the answer instruction. The egocentric-only baseline uses only \texttt{Exp1}, whereas \ourmodel uses both \texttt{Exp1} and \texttt{Exp2}.

\begin{figure}[!t]
    \centering
    \small
    \begin{tcolorbox}[width=\linewidth, sharp corners=all, colback=gray!10, boxrule=0.3mm]
    \ttfamily\setlength{\parskip}{0.5ex}

    {<|im\_start|>system} \\
    {\{system prompt\}<|im\_end|>} \\
    {<|im\_start|>user} \\
    {\textbf{Audio:}} \\
    {<|audio\_start|>{<|audio\_pad|>}<|audio\_end|>} \\[1ex]

    {\textbf{Video 1:  \# Egocentric}} \\
    {<|vision\_start|>{<|video\_pad|>}<|vision\_end|>} \\[1ex]

    {\textbf{Video 2:  \# Floormap}} \\
    {<|vision\_start|>{<|video\_pad|>}<|vision\_end|>} \\[1ex]

    {You are an expert AI assistant analyzing two temporally synchronized videos and audio. \{ego/exo perspective\}} \\
    {\textbf{The videos show two concurrent views:}} \\
    \hspace*{0.5em}{- \textbf{\textcolor{Red}{Video 1: Camera wearer's view}}} \\
    \hspace*{0.5em}{- \textbf{\textcolor{Blue}{Video 2: Top-down floormap}}} \\
    \hspace*{1.0em}{- \textbf{Grid axes: Distance in meters (1 unit = 1m)}} \\
    \hspace*{1.0em}{- \textbf{Named objects/landmarks placed at their estimated positions}} \\
    \hspace*{1.0em}{- \textbf{Red dot/arrow: Camera position/heading}} \\
    \hspace*{1.0em}{- \textbf{Yellow area: Camera FOV}} \\
    \hspace*{1.0em}{- \textbf{Green dot: Sound source}} \\[1ex]

    {Answering Strategy:} \\
    {1. Identify the audio event time.} \\
    {2. Check Video 1 at that time as your primary reference. Use this view as a main perspective for answering.} \\
    {3. If spatial details are uncertain, read the grid coordinates on Video 2 to estimate distances and relative directions between relevant markers.} \\[1ex]

    {Question:} \\
    {\{question\}} \\[1ex]

    {\{options, if any\}} \\[1ex]

    {\{answer formatting\}<|im\_end|>} \\
    {<|im\_start|>assistant}

    \end{tcolorbox}
    \vspace{-10pt}
    \caption{\textbf{Instruction prompt of our \ourmodel framework for Qwen-Omni models.} Gemini models use similar prompt with their own system prompt and multimodal token sequence.}
    \label{fig:sample_prompt}
\end{figure}

\noindent\textbf{Video 1: Egocentric view (Exp1).} This explanation identifies the camera wearer's first-person view and its synchronized audio, clarifies the reference perspective, and directs the model to locate the queried sound event and relevant visual evidence.

\begin{tcolorbox}[breakable, width=\linewidth, sharp corners=all, colback=gray!10, boxrule=0.3mm, top=1mm, bottom=1mm, left=1mm, right=1mm]
\footnotesize\ttfamily\setlength{\parskip}{0.35ex}
{\textbf{Exp1:}} \\
Video 1: [Your / The camera wearer's] egocentric view. \\
- This first-person video, with synchronized scene audio, shows what the camera wearer sees and hears. \\
- For an exocentric task, the robot in the question and the camera wearer are distinct entities; answer from the robot's perspective. \\
- Use Video 1 to pinpoint the target sound, check direct visibility, and identify visually described objects. \\[0.5ex]
\end{tcolorbox}

\noindent\textbf{Video 2: Floormap view (Exp2).} For \ourmodel, we additionally provide an explicit description of the rendering context to guide interpretation of the proposed 2D floormap. This explains the metric grid, camera position and heading, field of view, object landmarks, and estimated sound-source markers. These definitions clarify how to interpret the rendered spatial information and help align it with the corresponding egocentric observations at the same time, allowing the model to jointly use visual evidence and map geometry.

\begin{tcolorbox}[breakable, width=\linewidth, sharp corners=all, colback=gray!10, boxrule=0.3mm, top=1mm, bottom=1mm, left=1mm, right=1mm]
\footnotesize\ttfamily\setlength{\parskip}{0.35ex}
{\textbf{Exp2:}} \\
Video 2: Top-down floormap synchronized with Video 1. \\
- Grid axes show distance in meters (1 cell = 1 m). \\
- Red dot and arrow: the camera wearer's position and heading. Yellow wedge: camera field of view. \\
- Small colored dots mark detected object positions; displaced bold labels name the objects. \\
- Green dot: an approximate active sound-source estimate, smoothed over $\pm 1$ s; it can be inaccurate or absent. \\
- The floormap does not show the other person's facing direction. A clearly visible target is best judged from Video 1. \\[0.5ex]
\end{tcolorbox}

The templates above describe the shared video context; minor guidance variants account for each task's reference perspective and reasoning requirements.

\section{Limitations and Future Work}
\label{app:limitations}

\paragraph{Dependency on high-fidelity sensor data.}
The geometric fidelity of our rendered 2D floormaps is inherently bounded by the high-quality sensor suite used in the AEA dataset. Specifically, the construction of our accurate spatial prompts relies heavily on clean 3D point clouds, sub-centimeter camera poses, and multi-channel spatial audio. While our pipeline effectively localizes sound sources, acoustic direction and distance estimation remains inherently noisy in highly reverberant environments or scenes with overlapping audio sources.
This indicates that as foundational datasets like AEA expand and wearable sensor technologies continue to improve, the precision of our rendered spatial hints will naturally increase, providing substantial room for further performance advancements within our framework.

\paragraph{Lack of cognitive abilities for plot/grid style.}
Current foundation AV-LLMs are predominantly trained on vast corpora of natural images and standard video formats. Therefore, explicitly injecting top-down metric maps, characterized by grid lines and abstract geometric plots, introduces a severe out-of-distribution visual input challenge. As observed in Appendix\,\ref{app:qualitative}, the models often struggle to accurately read and identify precise metric coordinates, or cognitively relate the global coordinate frame back to the first-person egocentric perspective. Although optionally incorporating semantic grounding with explicit object labels partially mitigates this issue, this core representational mismatch between natural visual data and abstract geometric maps remains a fundamental cognitive bottleneck for current architectures. We also believe that our framework can scale with the model's intrinsic map reading capabilities.

\paragraph{Temporal localization dependency.}
Answering spatial queries in SAVVY-Bench~\cite{chen2025savvy} and \ourbench requires precise temporal grounding of sound events within continuous video streams. Errors in identifying the query moment or interval can lead the model to reason over the wrong spatial context, even when its geometric reasoning is otherwise correct. Performance therefore reflects both temporal localization and spatial reasoning, making it difficult to isolate the latter. Beyond this evaluation dependency, requiring temporal grounding in the audio track as a prerequisite inherently restricts task design. Future evaluation should both disentangle temporal localization errors from downstream spatial reasoning errors and extend task formulations beyond sound event based temporal grounding.

\paragraph{Integrated reasoning across categories.}
\ourbench broadens spatial evaluation through three complementary categories: \textit{dynamic relativity}, \textit{regional}, and \textit{path reasoning} QAs. However, the current tasks assess reasoning within each category independently rather than requiring models to integrate capabilities across categories. Real-world interactions may demand such integration, combining cross-agent viewpoint transformation, semantic region understanding, and recognition of physical surroundings within a single task. Since \ourbench does not explicitly evaluate these complex scenarios, extending it with tasks that require integrated reasoning across categories could bring spatial evaluation into closer alignment with real-world agent interactions.

\end{document}